\documentclass[a4paper,fleqn]{cas-sc}

\usepackage[numbers]{natbib}

\begin{document}
\let\WriteBookmarks\relax
\def\floatpagepagefraction{1}
\def\textpagefraction{.001}
\shorttitle{Short-term forecasting of global solar irradiance with incomplete data}
\shortauthors{Laura S. Hoyos-Gómez et~al.}

\title [mode = title]{Short-term forecasting of global solar irradiance with incomplete data}                      



\author[1]{Laura S. Hoyos-Gómez} [type=editor,
                        orcid=0000-0001-7511-2910]
\cormark[1]
\ead{lshoyosg@unal.edu.co}
\credit{Writing - Original draft preparation, Methodology, Software, Formal analysis, Validation}

\address[1]{Grupo de Investigación en Potencia Energía y Mercados,Universidad Nacional de Colombia, Manizales, Caldas, Colombia}

\author[2]{Jose F. Ruiz-Muñoz}
\ead{jruizmunoz@ufl.edu}
\credit{Methodology, Software, Formal analysis, Writing - review \& editing, Formal analysis}

\author[1]{Belizza J. Ruiz-Mendoza} [%
   orcid=0000-0003-3016-7787]
\ead{bjruizm@unal.edu.co}
\credit{Funding acquisition, Writing - review \& editing, Supervision}

\address[2]{Department of Electrical and Computer Engineering, University
of Florida, Gainesville, Florida, USA}

\cortext[cor1]{Corresponding author}


\begin{abstract}
Accurate mechanisms for forecasting solar irradiance and insolation provide important information for the planning of renewable energy and agriculture projects as well as for environmental and socio-economical studies. This research introduces a pipeline for the one-day ahead forecasting of solar irradiance and insolation that only requires solar irradiance historical data for training. Furthermore, our approach is able to deal with missing data since it includes a data imputation state. In the prediction stage, we consider four data-driven approaches: Autoregressive Integrated Moving Average (ARIMA), Single Layer Feed Forward Network (SL-FNN), Multiple Layer Feed Forward Network (FL-FNN), and Long Short-Term Memory (LSTM). The experiments are performed in a real-world dataset collected with 12 Automatic Weather Stations (AWS) located in the Nariño - Colombia. The results show that the neural network-based models outperform ARIMA in most cases. Furthermore, LSTM exhibits better performance in cloudy environments (where more randomness is expected).
\end{abstract}

\begin{graphicalabstract}
\includegraphics[width=\textwidth]{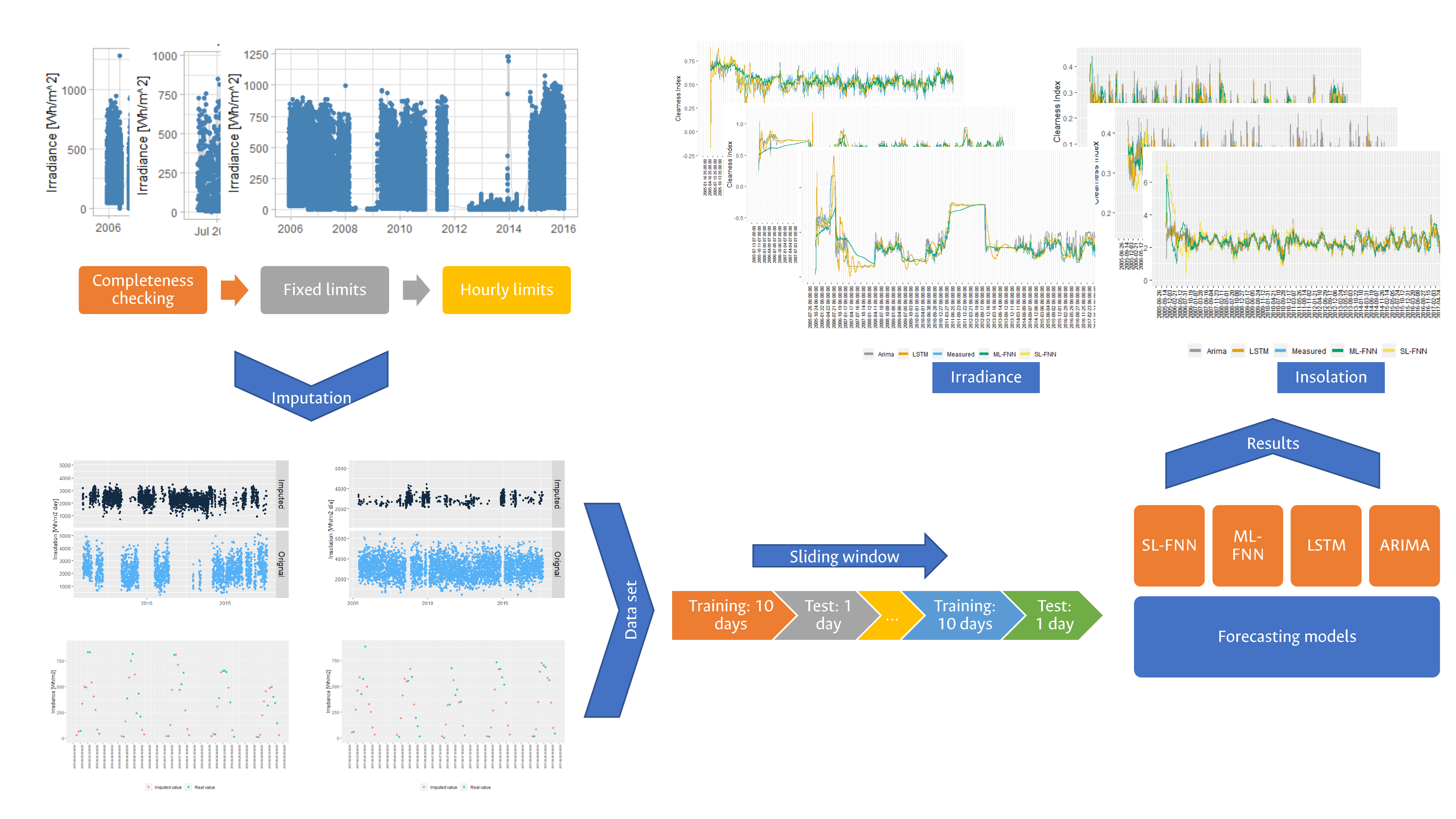}
\end{graphicalabstract}

\begin{highlights}
\item Forecasting the clear sky index facilitates the use of solar energy on electricity generation, agriculture, and ecology studies.
\item We proposed a framework that uses linear and non-linear data-driven models for the one-day ahead forecasting of clear sky index in a tropical environment.
\item We implement imputation techniques for deal with incomplete time series.
\item Our results show that the forecasting performance depends on the environmental conditions and amount of training data.
\end{highlights}

\maketitle

\section{Introduction}\label{S1}


The proliferation of solar power generation systems has promoted the interest in solar irradiance and insolation forecasting models. Accurate forecasting systems allow estimating the electricity generation in the long- medium- and short-term. This information is crucial to maintain the balance between energy demand and supply \citep{Dannecker2015}, as well as minimize costs associated with start and shutdown of conventional power plants \citep{Badosa2017}. 

\subsection{Background}
 
Data recorded in extended periods of time has been used to understand the behavior and predict future solar irradiance values in a specific location using time series analysis \citep{Shumway2011, Suehrcke2000}. The solar irradiance is a time-dependent phenomenon composed of a deterministic and stochastic part \citep{Boland2008}. Mathematical models allow predicting the exact future value of the deterministic part. On the other hand, the stochastic part outputs a future value between two limits with a confidence interval \citep{Box2016}. The forecasting accuracy relies on the stochastic component ability to model the solar irradiance changes induced by the clouds  \citep{Inman2013a}. There are several approaches to forecast the solar radiation such as persistence methods that assume that the value at time $t+1$ is equal to value at time \(t\) \citep{Diagne2013}, autoregressive models that allow modeling stationary and non-stationary variations and describing complex nonlinear atmospheric phenomena \citep{Inman2013a}, e.g., autoregressive moving average (ARMA), and autoregressive integrated moving average (ARIMA); and soft computing techniques, e.g., support vector machine (SVM), artificial neural network (ANN), and fuzzy and genetic algorithms (GA) \citep{Demirhan2018}. The ANN, fuzzy logic, and hybrids are robust for modeling the physical processes' stochastic nature, like the solar irradiance because of their capacity to compensate systematic errors and problematic learnable deviation \citep{Paulescu2013,KUMARI2021117061}.

The autoregressive models describe the characteristics and behavior of the time series using an autoregression process \citep{Antonanzas2016,Dannecker2015}. The motivation for using these forecasting models is their robustness to random errors and outliers \citep{Diagne2013,Sobri2018}. An ANN is a statistical model that establishes a relationship between the input and output data during a training process through layers formed by interconnected nodes of inputs, outputs, hidden layers, and activation functions. It has become one of the most popular solar power forecasting technique \citep{Mazorra-Aguiar2018,Antonanzas2016}. Long Short-Term Memory (LSTM) network is an advanced Recurrent Neural Network (RNN), which has been recently used in the renewable energy field  \citep{Chandola2020}. LSTM learns the dependence between successive data \citep{Ghimire2019}. Some studies suggest that LSTM outperforms other state-of-the-art models in forecasting day-ahead solar irradiance \citep{Husein2019}. The selection of a forecasting method depends on the desired timescale, e.g., intra-hour (15 min to 2 h), intra-day (1h to 6h) and day ahead (1 day to 3 days) \citep{Diagne2013}. Statistical approaches usually perform well for short-term forecasting, such as ARIMA and ANN. For long-term analysis, soft computing techniques are frequently preferred  \citep{Coimbra2013,Demirhan2018}.

Each one of the literature surveys in \cite{Voyant2017,Sobri2018,Blaga2019,Guermoui2020} reviewed between 40 to 85 studies that implement different solar forecasting methods, temporal horizon, input variables, data periods and locations. We found that there is a scarcity of studies analyzing tropical environments since only 18 of more than 160 papers are focused in this geographical region. The studies analyzing tropical environments use as input satellite information, and in-situ meteorological variables such as temperature, precipitation, and global solar irradiance. For example, \citep{Wu2011} uses ARMA and Time Delay Neural Network (TDNN) to predict hourly Global Horizontal Irradiance (GHI) using as input the same variable in Singapore. \citep{Dong2013} forecasts hourly GHI with an exponential smoothing state space model (ESSS), ARIMA, linear exponential smoothing (LES), simple exponential smoothing (SES) and random walk (RW) using data from Singapore. \citep{Wu2013} segments the GHI in clusters according patterns with K-means algorithm, and TDNN predicts the GHI. \citep{Pan2013} estimates the monthly clearness index in 23 locations of India and uses a multi-gene genetic programming (MGGP) algorithm that have latitude, longitude, altitude, month of the year, temperature ratio and mean duration of sunshine per hour as input variable. \citep{Dong2014} forecasts hourly GHI with satellite images analysis and a hybrid ESSS together with ANN for Singapore. \citep{Wu2014} implements a genetic approach combining multi-model framework for predict GHI five minuted ahead in Singapore. \citep{Wu2014a} uses solar power output, GHI, air and module temperature to predict one hour ahead PV power output in Malaysia (the implemented prediction models were ARIMA, SVM, ANN, ANFIS and a combination models using GA algorithms). \citep{Yadav2014} decomposes the GHI in approximate and detail components with wavelength transform and then use a RNN to forecast hourly GHI in Rajasthan, India. \citep{Dong2015} forecasts hourly GHI for Singapore using a hybrid model that combines self-organizing maps, support vector regression and particle swarm optimization. This model has as input the GHI measure in a tropical environmental characterized by high variability due to the combination of temperature, humidity and evaporation that favor the cloud formation. \citep{Royer2016} implements an iterative combination of wavelet ANN (CWANN) to forecast the hourly GHI using as input the GHI measured in ten cities of Brazil. \cite{Mukaram2017} predicts the monthly GHI in three location of Malaysia with SARIMA, ANN and a combination of both models. \citep{Monjoly2017} forecasts the hourly GHI in the Guadeloupe island. They detrend the GHI using the clear sky index as input, and decompose the GHI to use the autoregressive and ANN models for forecasting.

Solar irradiance datasets frequently have a considerable amount of missing data. To overcome this problem, \cite{Layanun2017a} proposed a data-missing imputation technique for a seasonal ARIMA-based forecasting method in Bangkok in Thailand that averages the data classified by weather type determined with temperature and humidity information. Likewise, \cite{Rodriguez-Rivero2017} applied an average smoothing technique to fill the time series's missing data to short-term forecasting. \cite{Ogunsola2014a} uses three approaches to solar irradiance data imputation Singular Spectral Analysis (SSA), Statistically Adjusted Solar Radiation (SASR), and Temperature-based Approach (TBA).



\subsection{Contribution}
This study applies four state-of-the-art prediction models for global solar irradiance and insolation one day-ahead forecasting in tropical and mountainous environments with incomplete data: i) ARIMA, ii) Single Layer Feedforward Neural Network (SL-FNN), iii) Multi-Layer Feedforward Neural Network (ML-FNN), iv) and Long Short-Term Memory (LSTM).

The solar irradiance $[Wh/m^2]$ is the amount of solar energy on a specific area during a specific time interval, and solar insolation $[Wh/m^2 day]$ is the cumulative solar energy on a surface  \citep{sandianationallaboratories2021}.

In the irradiance case, the models forecast the hourly irradiance values one day ahead. For the insolation, which in this case is the sum of the irradiance hourly values, the models forecast the insolation one day ahead. The missing data imputation combines a linear interpolation of the subsequent values with the average of past values measured at the same hour of the imputed data for hourly data, and a TBA for daily data. We use the clear sky index to remove the deterministic effect. 

\subsection{Paper structure}
Section \ref{S2} contains the forecasting models, the quality control procedure, data imputation and model performance. Section \ref{S3} includes dataset description, data pre-processing, model forecasting model and training settings. Section \ref{S4} consists of three parts, the first one \ref{S4.1} with the imputation results, and \ref{S4.2} and \ref{S4.3} contain the hourly and daily forecasting results correspondingly. Section~\ref{S5} concludes the paper.

\section{Methods}\label{S2}

The solar irradiance and insolation can be computed when the clear sky index $(k_c)$ is known \citep{Dorvlo2002,ISLAM2009511}. Since $(k_c)$ estimates the clear irradiance without the clear atmosphere effect,  $(1-k_c)$ estimates the attenuation caused by cloudiness~\citep{Brownson2014}. The $k_c$ is the ratio between the global solar irradiance at ground level and the clear sky global solar irradiance, computed with the Kreith \& Kreider model \citep{Sen2008}. 

\begin{figure}[htb!]
    \centering
    \includegraphics[width=0.8\linewidth]{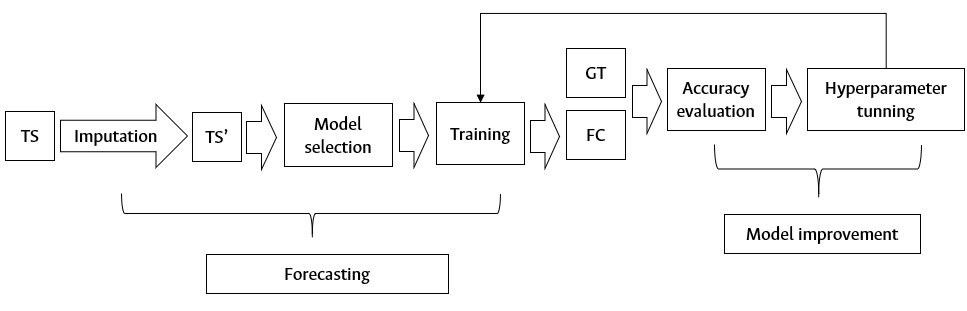}
    \caption{Forecasting models and hyperparameter tuning stages. TS, TS', FC, and GT stand for input time series, imputed time series, forecasting (model output), and ground-truth (value provided by the AWS), respectively.}
    \label{fig:met4}
\end{figure}

Figure~\ref{fig:met4} describes our forecasting methodology. As a preprocessing step, we perform ``quality control'' on the input data by removing anomalous data. After that, we fill the time series gaps (missing data). In the model selection step, we consider four options: ARIMA, SL-FNN, ML-FNN and LSTM. In the training step, the model's parameters are computed in such a way that fit the training data. To evaluate the model, we use the parameter values updated until the $t$-th day to predict the irradiance or insolation in the $t+1$-th day. We use the performance measures to tune the hyperparameters. 

\subsection{ARIMA}\label{S2.1}
ARIMA consists of three components: i) autoregression (AR), ii) integration (I), and iii) moving average (MA). AR and MA deal with the stochastic elements, and $I$ renders the time series stationarity. ARIMA is denoted as ARIMA \(\left(p,d,q\right)\), where $p$ and $q$ are the AR and MA order, respectively, and $d$ is the number of derivatives applied to the time series (e.g., $d=0$ means that the time series is already stationary, and $d=2$ means that two derivatives are needed to make the data stationary) \citep{AgamiReddy2011}. ARIMA \(\left(p,d,q\right)\) is defined by:
\begin{equation}
\label{eq:arima}
\begin{matrix}
\Phi _{p}\left ( B \right )\Delta ^{d}x_{t}=\theta _{q}\left ( B \right )u_{t},&  u_{t}\sim WN\left ( 0,\sigma ^{2} \right ) 
\end{matrix}
\end{equation}
or 
\begin{equation}
\label{eq:arima2}
    \left ( 1-\sum_{i=1}^{p}\Phi _{i}B^{i} \right )\left ( 1-B \right )^{d}x_{t}=\left ( 1+\sum_{i=1}^{q} \theta _{i}B^{q}\right )u_{t}
\end{equation}
where $B$ is the backward shift operator that may be treat as a complex number, \(\Delta=1-B\) is the backward difference, \(\Phi_{p}\) and \(\theta_{q}\) are polynomials of order $p$ and $q$ respectively. \(\left(1-B\right)^{d} x_t\) is the responsible to transform the non-stationary time series in a stationary one. After applying this transformation, it is possible to use any forecasting strategy for stationary data \citep{Montgomery2008,Shumway2011}.

AR describes the past behavior of the time series and series residual at the actual time as a weighted linear combination of values of a dataset of a stochastic process  \(x_t\) \citep{Dannecker2015} and a white noise \(u_t\) as follows
\begin{equation}
\label{eq:ar1}
    x_{t}=\Phi _{1}x_{t-1}+\Phi_{2}x_{t-2}+\cdots +\Phi _{p}x_{t-p}+u_{t}=\sum_{i=1}^{p}\Phi_{i}x_{t-i}+u_{t}.
\end{equation}

Using the backshift operator the equation \ref{eq:ar1} is expressed as follows:
\begin{equation}
    \label{eq:ar2}
    u_{t}=\left(1- \sum_{i=1}^{p}\Phi_{i}B^{i}\right)x_t.
\end{equation}

MA describes the time-series’ random perturbations by a weighted linear combination of previous values of a white noise error. Thus, the time series is represented as a set of uncorrelated and normal-distributed random variables, as follows:
\begin{equation}
    \label{eq:ma1}
    x_{t}=\theta _{1}u_{t-1}+\theta_{2}u_{t-2}+\cdots +\theta _{q}u_{t-q}=\left(1-\sum_{i=1}^{p}\theta_{i}B^{q}\right)u_{t}.
\end{equation}

\subsection{Feedforward Neural Networks (FNN)}\label{S2.2}

An FNN is an ensemble of units, known as neurons, connected by synaptic joints each one with a weight coefficient \citep{Blaga2019}. An FNN can be divided into three parts: input, hidden layers, and output. The first part receives the input data. The hidden layers connect the input and output. The output layer outcomes the computed values \citep{Premalatha2016a}.

The FNN training requires an iterative backpropagation procedure that learns an input-output mapping. This process has four steps: i) forward propagation of the training pattern input, ii) error calculation by a loss function that compares estimated and reference values, iii) backpropagation of the error to recompute each weight \(\Delta w_{ij}\) from the output to the hidden layer, and iv) weight's \(w_{ij}\) updating: \(w_{ij}^{new}=w_{ij}^{old}+\lambda \Delta w_{ij}\), where \(\lambda\) is known as learning rate \citep{Blaga2019}.

During forward propagation, the inputs $\textbf{x}^{i-1}$ are multiplied by the weights $w$, the individual results are summed-up, and a bias $b$ is added to the results as an offset value as follows
\[
    z^{(i)}_n =\sum^{N_{i-1}}_{k=1} w^{(i)}_{nk} x^{i-1}_k +b^{(i)}_n
\]
where $i$ is the current layer (if $i=1$, $\textbf{x}^0$ is the input), $n$ is the $n$-th neuron in the current layer, $k$ is the $k$-th neuron in the previous layer, and $N_{i-1}$ is the number of neurons in the previous layer. This result is passed through a so-called activation function \citep{Ghimire2019a} such that
\[
    \textbf{x}^{(i)}=f(\textbf{z}^{(i)})
\]
where $\textbf{x}^i$ is the current layer's output, and $f(\cdot)$ is typically used to bound the signal or induce non-linear interactions. In this way, the signal passes all the layers until reaching the output. The weights are initialized with random values. Both weights and bias are iteratively updated until a stopping criterion is fulfilled \citep{Munawar2020}.

\begin{figure}[htb!]
    \begin{subfigure}[b]{0.285\textwidth}
    \centering
    \includegraphics[width=\textwidth]{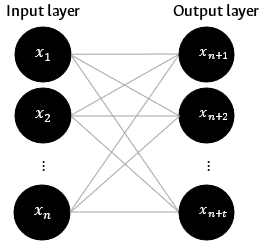}
    \label{fig:slp}
    \end{subfigure}
    ~
    \begin{subfigure}[b]{0.45\textwidth}
    \centering
    \includegraphics[width=\textwidth]{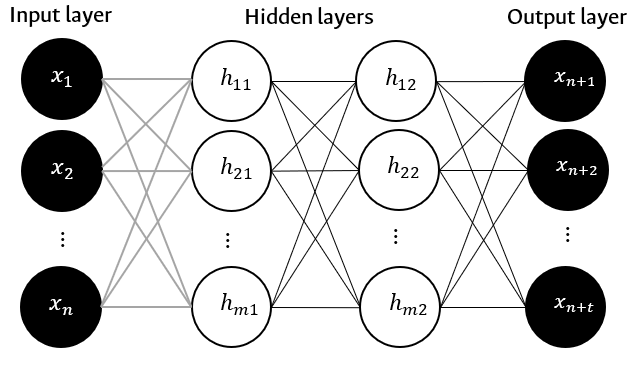}
    \label{fig:mlp}
    \end{subfigure}
\caption{FNN architectures used for time series forecasting. (a) SL-FNN: computes the prediction by a linear combination of the input data. (b) ML-FNN: FNN with hidden layers and non-linear activations for learning non-linear mapping.}
\label{fig:anns}
\end{figure}

In this study, we consider two FNN architectures: a single layer FNN (SL-FNN) that links directly input and outputs without activation functions, and a multi-layer FNN (ML-FNN) with two hidden layers and ReLU activation functions between them. Note that our SL-NN and ML-NN architectures learn a linear and non-linear mapping, respectively. Figure~\ref{fig:anns} illustrates both of them.

\subsection{LSTM}\label{S2.3}
In addition to the FNN models described above, we consider a recurrent neural network (RNN) architecture. RNN is an efficient tool to deal with temporal patterns due to the capacity of remember previous data \citep{Chandola2020}. However, training a RNN is usually difficult due to the vanishing gradient problem~\citep{Husein2019}. Given an input \(x_{t-1}=x_1,x_2,x_n\), the output \(x_t\) in a RNN is given by: 
\begin{equation}
    \label{eq:lstm}
    \begin{matrix}
    x_{t}=w_{hx_t}h_{t}+b_{x_t}\\
    h_{t}=H(w_{hx_{t-1}}x_{t-1}+w_{hh}h_{t-1}+b_{h})
    \end{matrix}
\end{equation}
where \(w_{hx_{t-1}}\), \(w_{hh}\), \(w_{hx_t}\) are input-hidden, hidden-hidden and hidden-output weight matrices, \(b_h\), and \(b_{x_t}\) are hidden and output bias vectors, respectively. $H$ term is the hidden layer activation function.

\begin{figure}[htb!]
\centering
\includegraphics[width=0.75\linewidth]{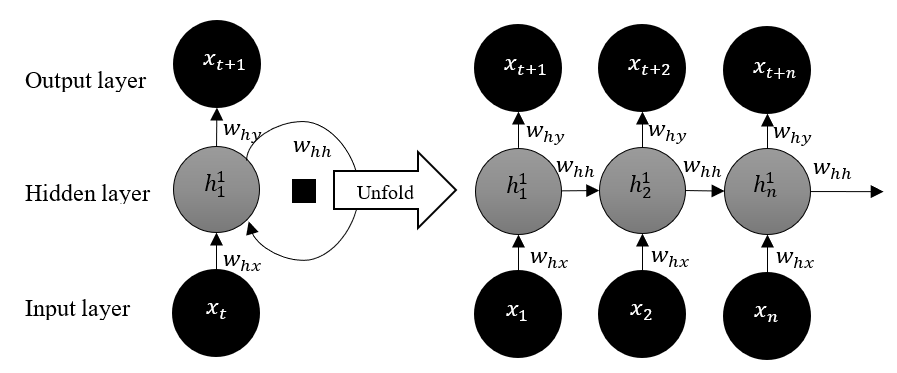}
\caption{A RNN structure learns sequential patterns.}
\label{fig:rnn}
\end{figure}

\begin{figure}[ht]
\centering
\includegraphics[width=0.65\linewidth]{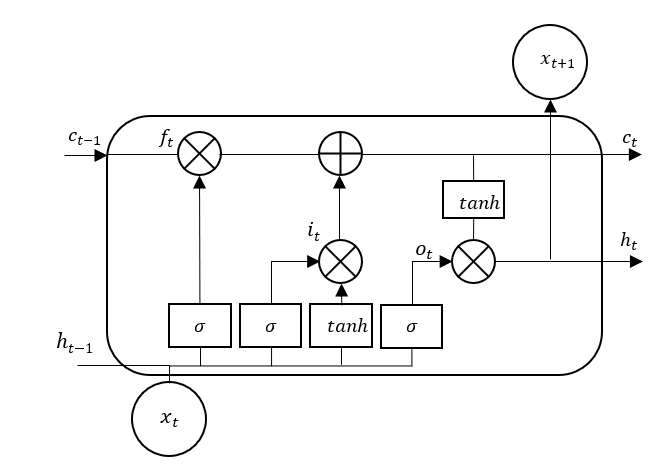}
\caption{An LSTM is an RNN with an explicit memory unit.}
\label{fig:lstm}
\end{figure}

LSTM is an advanced RNN that resolves the gradient problem including an explicit memory to the network. LSTM has an input, forget gate, output gate, and a cell unit that serves as memory for a defined time interval (see Figure \ref{fig:rnn}). The gates control the flow of information that enter and leave the cell unit, see Figure \ref{fig:lstm} \citep{Husein2019}. The forget gate \(f_t\) determine the influence of the previous state on the current state. The input gate \(i_t\) receives the new information to update the cell state. The output gate \(o_t\) provide the information based on the cell state. The sigmoid function \(\sigma\) adjust the output values of these gates to a value between 0 to 1, that are interpreted as a probability \citep{Kwon2020}. 

The explicit memory of the LSTM makes this technique appropriate for a short-term, near real-time forecast model \citep{Ghimire2019}.  The LSTM unit has three operation states: 1)~Input gate activate: the cell memory accumulates new information. 2)~Forget gate activate: the cell erases the accumulated information. 3)~Output gate activate: the final output propagates to the ultimate state.

In the LSTM forward propagation the forget gate computes
\begin{equation}
    \label{eq:lstm1}
    f_{t}=\sigma(w_{xf}x_{t}+w_{hf}h_{t-1}+b_{f})
\end{equation}
where \(h_{t-1}\) represents the last previous state, and \(\sigma(\cdot)\) is the logistic sigmoid function.  The LSTM cell internal status is updated with a conditional self-loop weight \(f_{t}\) as follows:
\begin{equation}
    \label{eq:lstm2}
    c_{t}=f_{t}c_{t-1}+i_{t}\tanh(w_{xc}x_{t}+w_{hc}h_{t-1}+b_{c}).
\end{equation}
The cell unit \(c_t\) is a linear self-loop controlled by the forget gate unit \(f_t\) that determines the forward contribution of \(c_{t-1}\). Then, the external input unit \(i_t\) is calculated similarly to the forget gate with its own parameters:
\begin{equation}
    \label{eq:lstm3}
    i_t=\sigma(w_{xi}x_{t}+w_{hi}h_{t-1}+b_{i}).
\end{equation}
The output \(h_t\) of the LSTM cell can also be shut off, via the output gate \(o_t\), which also uses a sigmoid unit for gating:
\begin{equation}
    \label{eq:lstm4}
    h_t=o_{t}\tanh(c_{t})
\end{equation}
where $ o_t=\sigma(w_{xo}x_{t}+w_{ho}h_{t-1}+b_o)$, and \(i,f,o\) and \(c\) are the input, forget, output gate and cell, respectively. Note that \(i,f,o\) and \(c\) have the same size as the hidden vector \(h\).

\subsection{Quality Control}\label{S2.4}
To reduce the risk of biases induced by inaccurate data, we apply a quality control process that consist of three stages: i)~Data completeness checking: samples that lack essential information (station code, variable code, date and time, and data value) are removed from the data set. ii)~anomaly detection based on fixed limits: to avoid over estimate the resource, we discard the samples that exceed the maximum hourly extraterrestrial solar irradiance value ($I_0$). Therefore, it is required that \(I_0\geq I_{mt} \) 
where \(I_{mt}\) is the measured global solar irradiance at time $t$, and $I_0$ is
\begin{equation}
\label{eq:isc1}
    I_{sc}\left [ 1+0,033cos\left ( 360\frac{D-3}{365} \right ) \right ]*sin\beta
\end{equation}
where \(D\) is the Julian day, \(I_{sc}\) is the solar constant \(\left(1.367 \left [ W/m^{2} \right ] \right)\) representing the energy from the Sun per unit area of the surface perpendicular to the irradiance propagation direction \citep{Sen2008} and \(sin\beta=cos\phi cos\delta cos\omega_{s}+sin\phi sin\delta.\)  iii)~anomaly detection based on hourly limits: to avoid underestimate the resource, we use as lower bound the 3\% of the clear sky global solar irradiance ($I_{cst}$), where $I_{cst}=I_{0}\tau$. We consider the samples under the lower bound as anomalous values and remove them from the dataset.

To estimate the atmospheric transmittance $\tau$ and \(I_{cst}\), we implement the Kreith \& Kreider model \citep{Sen2008}:
\begin{equation}
\label{eq:kkcsm1}
    \tau=0,56\left(e^{-0,65/\sin \beta}+e^{-0,095/\sin \beta}. \right)
\end{equation}

The procedure above follows the UNE500540 regulation~\citep{AENOR2004}, and the recommendations in \citep{Estevez2011}, which applied quality control to solar irradiance data in Spain. 

\subsection{Data imputation}\label{S2.5}
Data imputation is the process of completing the missing data with reasonable values. In this case, the authors use the clear sky index $k_c$ instead of dealing with the solar irradiance data directly.  

In our experiments, for hourly data we filled out the missing data on the first day with a value of 1. Afterward, the technique considers three cases: if the missing data is at i) 6, ii) 18, or iii) between 7 to 17 hours. In the first case, the missing data is the average between the value at the same hour of the day before with the clear sky index value of the 7 hours of the current day. On the second case, the value is the result of the average between the data at 17 hours of the current day and the value at 18 hours of the day before. In the third case, the imputed value corresponds to the average between the value at the same hour the day before and the value of the immediately previous and following hours. If the immediately next value is missed, it is not considered in the calculation.

For daily data imputation, we used empirical temperature-based models such as Hargreaves and Samani \citep{Hargreaves1982},  

\begin{equation}
    \label{eq:hsm}
    \frac{H}{H_{0}}=a\left(T_{max}-T_{min}\right)^{0,5}
\end{equation}

where $H$ is the daily global solar insolation, $H_0$ is the daily extraterrestrial global solar insolation, $a$ is an empirical coefficient computed from each dataset, $T_{max}$ is the maximum daily temperature, and $T_{min}$ is the daily minimum temperature, and a new approach based on logistic regression,  

\begin{equation}
    \label{eq:logpro}
    \frac{H}{H_0}=\frac{1}{1+e^{-\left(a+b(T_{max} -T_{min})\right)}}
\end{equation}

where $a$ and $b$ are empirical coefficient computed from each dataset. These models estimate the solar insolation from the difference between the daily maximum and minimum temperature.

\subsection{Model performance estimation}\label{S2.6}
We use statistical validation to measure the forecasting performance. Table \ref{tab:estaerr4} shows the computed errors. For MAE and RMSE, the lower the better, and for MBE, the closer to zero the better \citep{Abreu2018}. The performance evaluation only considers the measure data, since the imputed data is used exclusively for training. Therefore, the amount of days for the error estimation is different from each AWS's total amount of data. 

\fontsize{9}{11}\selectfont{
\begin{longtable}[c]{@{}p{2.5cm}p{9cm}l@{}}
\caption{Statistical errors}
\label{tab:estaerr4}\\
\toprule
Measurement & Definition & Formula$^{*}$ \\* \midrule
\endfirsthead
\multicolumn{3}{c}%
{{\bfseries Table \thetable\ continued from previous page}} \\
\toprule
Measurement & Definition & Formula$^{*}$ \\* \midrule
\endhead
\bottomrule
\endfoot
\endlastfoot
Mean absolute error (MAE) & It is the average vertical distance between each predicted and observed point. This measure quantifies the error with more emphasis on the mean and less on individual extreme events.  & \(\frac{1}{n}\sum_{i=1}^{n}\left | p_{i}-o_{i} \right |\) \\
Root mean square error (RMSE) & It provides a measure of the error size and is sensitive to outlier values because this measure gives much weight on large errors. It captures variability rather than the overall trend. & \(\left [ \frac{1}{n}\sum_{i=1}^{n}\left ( p_{i}-o_{i} \right )^{2} \right ]^{1 \slash 2}\) \\
Mean bias error   (MBE) & This measure provides information on the long-term performance of the model, when the model has a systematic error that presents overestimated or underestimated predictors. Low values of MBE are desirable, though it should be noted that an overestimated data set will cancel another underestimated data set. & \(\frac{1}{n}\sum_{i=1}^{n} (p_{i}-o_{i}) \) \\
\bottomrule
\multicolumn{3}{l}{\begin{tabular}[c]{@{}l@{}}$^{*}$ \(p_{i}\) is the predicted value,\(o_{i}\) is the observed value, \(n\) is the amount of data\end{tabular}}\\
\caption*{Source: \citep{Almorox2011,DosSantos2014,Gueymard2014,Li2011,Mayer1993,Blaga2019}}
\end{longtable}
}
\normalsize

\section{Experimental set-up}\label{S3}

\subsection{Location and dataset}\label{3.1}
\begin{figure}[htb!]
\centering
\includegraphics[width=0.6\linewidth]{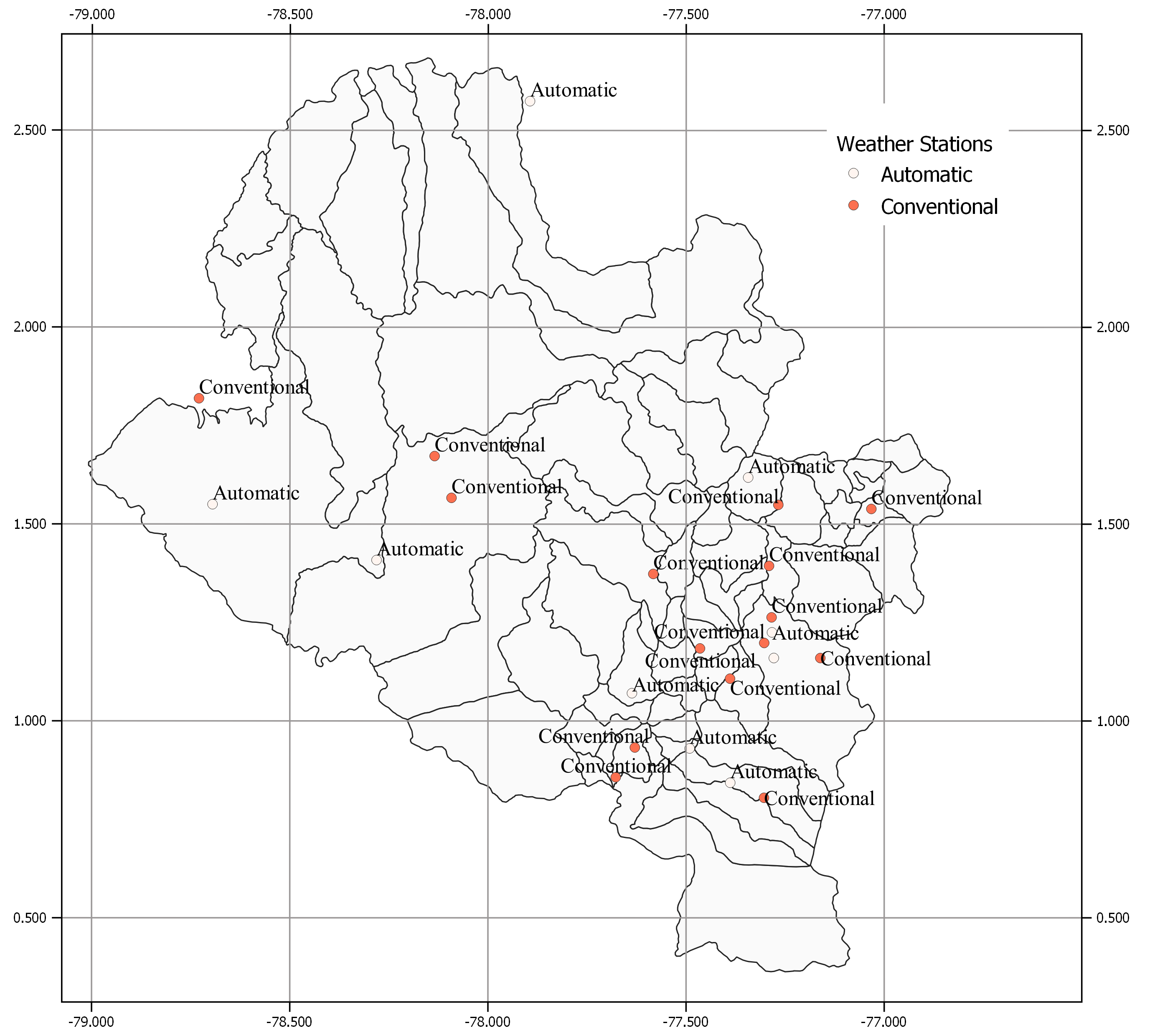}
\caption{Weather Stations Location}
\label{fig:aws3}
\end{figure}

In this study, we use irradiance data from twelve AWS (see Table \ref{tab:AWS1}) located in Nariño, Colombia, as shown in Figure \ref{fig:aws3}. This region is located in the Intertropical Convergence Zone, where the Andean mountain range splits into two mountain ranges. This zone is formed by three sub-regions, the Pacific, Andean, and Amazonia. The altitude of all the AWS ranges from 42 to 3.577 MASL. These geographical characteristics allow assessing the studied forecasting techniques on different physio-graphic and environmental conditions. 

\begin{table}[htb!]
    \centering
    \caption{Automatic Weather Stations}
    \resizebox{0.65\textwidth}{!}{%
    \begin{tabular}{llllll}
    \toprule
        AWS Name & Latitude& Longitude& Altitude& Period& Region \\ \midrule
        Biotopo & 1,41& -78,28& 512& 2005-2017& Pacific\\
        Altaquer& 1,56& -79,09&1.101& 2013-2014& Pacific\\
        Granja el Mira& 1,55& -78,69& 16& 2016-2017& Pacific\\
        Guapi &2,57&-77,89&42&2005-2017&Pacific\\
        Cerro-Páramo& 0,84& -77,39& 3.577& 2005-2017& Amazonia\\
        La Josefina& 0,93& -77,48& 2.449& 2005-2017& Andean\\
        Viento Libre& 1,62& -77,34& 1.005& 2005-2017& Andean\\
        Universidad de Nari\~{n}o& 1,23& -77,28& 2.626& 2005-2017& Andean\\
        Botana& 1,16& -77,27& 2.820& 2005-2017& Andean\\
        El Paraiso & 1,07& -77,63& 3.120& 2005-2017& Andean\\
        Sandona & 1,30& -77,46&1.838& 2016-2017&Andean\\
        Ospina Perez&1,25&-77,48&1.619&2016-2017&Andean\\
        
        \bottomrule
    \end{tabular}
    }
    \label{tab:AWS1}
\end{table}

\begin{figure}[htb!]
     \centering
     \begin{subfigure}[b]{0.32\textwidth}
         \centering
    \includegraphics[width=\textwidth]{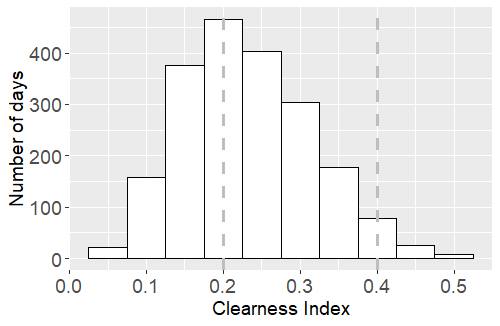}
         \caption{Biotopo}
         \label{biohis}
     \end{subfigure}
     \hfill
     \begin{subfigure}[b]{0.32\textwidth}
         \centering
         \includegraphics[width=\textwidth]{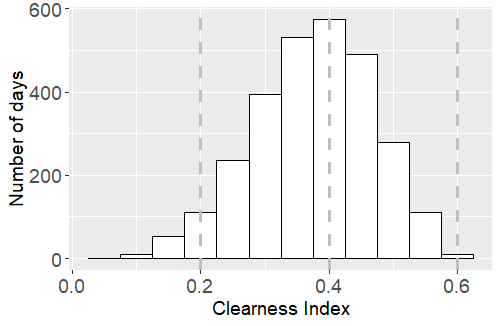}
         \caption{Viento Libre}
         \label{vlhis}
     \end{subfigure}
     \hfill
     \begin{subfigure}[b]{0.32\textwidth}
         \centering
         \includegraphics[width=\textwidth]{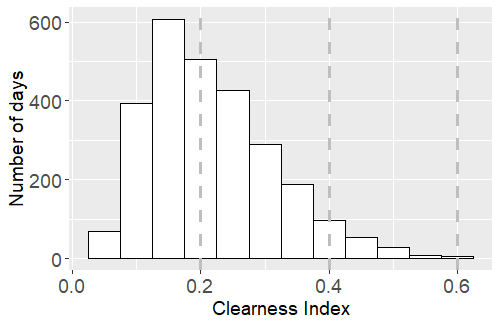}
         \caption{Cerro Páramo}
         \label{cphis}
    \end{subfigure} 
    \newline
     \begin{subfigure}[b]{0.32\textwidth}
         \centering
         \includegraphics[width=\textwidth]{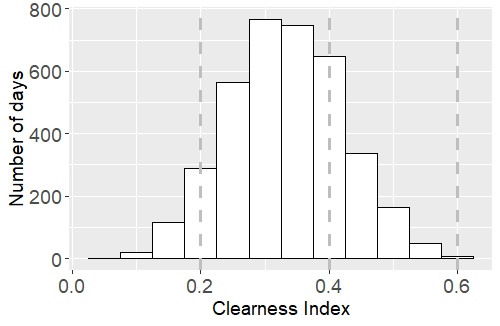}
         \caption{Universidad de Nariño}
         \label{udnarhis}
     \end{subfigure}
     \hfill
     \begin{subfigure}[b]{0.32\textwidth}
         \centering
         \includegraphics[width=\textwidth]{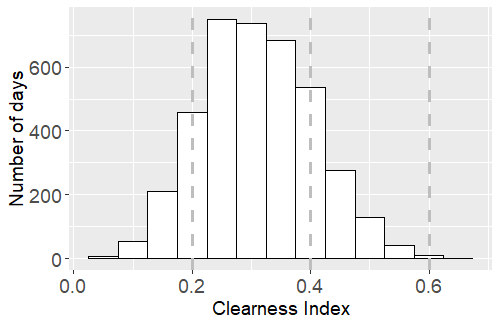}
         \caption{Botana}
         \label{bothis}
     \end{subfigure}
     \hfill
     \begin{subfigure}[b]{0.32\textwidth}
         \centering
         \includegraphics[width=\textwidth]{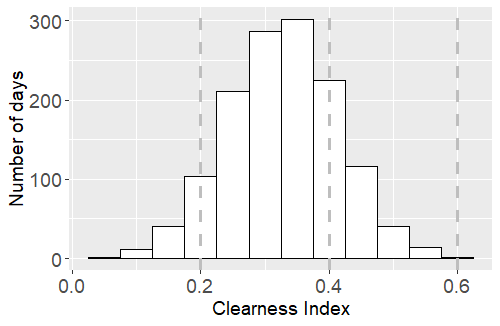}
         \caption{La Josefina}
         \label{jfhis}
     \end{subfigure}
    \newline
     \begin{subfigure}[b]{0.32\textwidth}
         \centering
         \includegraphics[width=\textwidth]{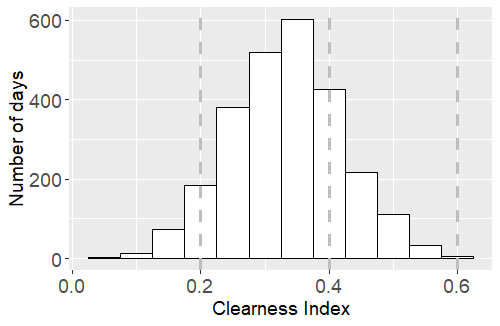}
         \caption{Paraiso}
         \label{prhis}
     \end{subfigure}
     \hfill
     \begin{subfigure}[b]{0.32\textwidth}
         \centering
         \includegraphics[width=\textwidth]{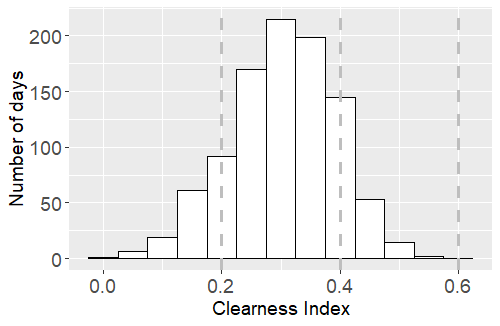}
         \caption{Guapi}
         \label{guhis}
     \end{subfigure}
     \hfill
     \begin{subfigure}[b]{0.32\textwidth}
         \centering
         \includegraphics[width=\textwidth]{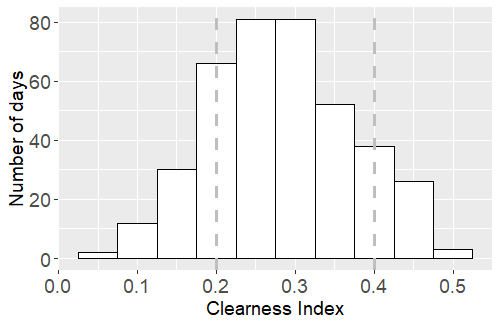}
         \caption{Altaquer}
         \label{alhis}
     \end{subfigure}
    \newline
     \begin{subfigure}[b]{0.32\textwidth}
         \centering
         \includegraphics[width=\textwidth]{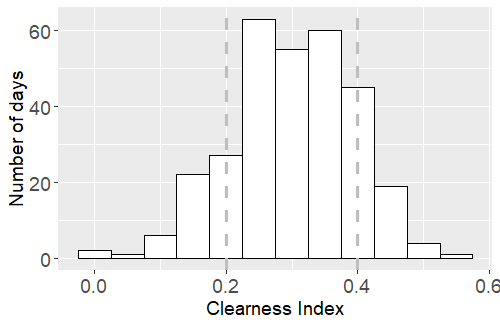}
         \caption{Granja el Mira}
         \label{gmhis}
     \end{subfigure}
     \hfill
     \begin{subfigure}[b]{0.32\textwidth}
         \centering
         \includegraphics[width=\textwidth]{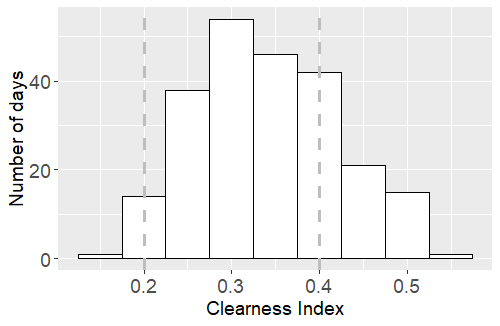}
         \caption{Ospina Perez}
         \label{ophis}
     \end{subfigure}
     \hfill
     \begin{subfigure}[b]{0.32\textwidth}
         \centering
         \includegraphics[width=\textwidth]{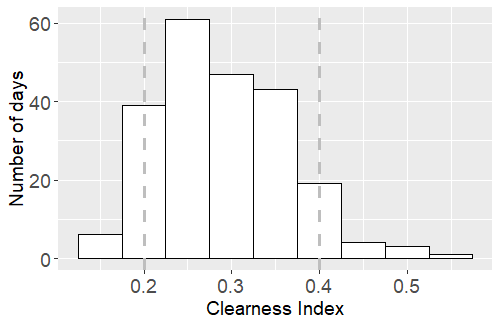}
         \caption{Sandona}
         \label{sanhis}
     \end{subfigure}
     \newline
     \caption{Days classification with clearness index: cloudy days \(0,0<K_t\leq0,2\), partially high cloudiness \(0,2<K_t\leq0,4\), partially low cloudiness \(0,4<K_t\leq0,6\), sunny \(0,6<K_t\leq0,75\), very sunny \(0,75<K_t\leq1,0\).}
    \label{fig:clinaws}
\end{figure}

Figure \ref{fig:clinaws} shows the clearness index $K_{t}$, which is the ration between the solar insolation and extraterrestrial solar insolation, based on the following categories \citep{Rivero2017}: cloudy days \(0,0<K_t\leq0,2\), partially high cloudiness \(0,2<K_t\leq0,4\), partially low cloudiness \(0,4<K_t\leq0,6\), sunny \(0,6<K_t\leq0,75\), very sunny \(0,75<K_t\leq1,0\). Note that \textit{Biotopo} and \textit{Cerro Páramo} have mostly cloudy days, and the others have mostly partially high cloudy days.

\subsection{Data pre-processing}\label{S3.2}
Forecasting methods aim to model the global solar irradiance's stochastic component. Therefore, in the pre-processing stage, we remove the time series' trend to reduce the deterministic effect. To this end, we use the clear sky index \citep{Benali2019,Diagne2013}. The $k_{c}$ ranges between 0 to 1 and indicates the amount of solar irradiance in a clear sky that reaches the Earth surface. Therefore, $k_{c}=0$ is obtained on an completely overcast day, while $k_{c}=1$ is achieved on a very sunny day.

\subsection{Forecasting models' architectures}\label{S3.3}
We use the forecasting models for the one-day ahead prediction of solar irradiance and insolation. We grouped the neural network architectures into SL-FNN, ML-FNN, and LSTM, and design two architectures per group (for irradiance and insolation forecasting in each case). Each architecture outputs a vector $\in \mathbb{R}^{13}$ and $\mathbb{R}$, respectively. In SL-FNN, the input directly feeds the output, so, the size of the input is 130 for solar irradiance forecasting (13 values per day), and 10 for insolation (1 value per day), and the output is 13 and 1, respectively. Our ML-FNN consists of two hidden layers with 130, and 10 neuron per hidden layer for irradiance and insolation, respectively, and ReLU activation functions between them. In LSTM, we use a a hidden layer (memory) with the same size as the input (130 or 10 accordingly) and a fully connected layer between this layer and the output. Note that there is not activation function that bounds the output in any of the neural network models.

\subsection{Training}\label{S3.4}
For training ARIMA, we use a sliding window that contains the clear sky index values of 10 days; therefore, the input range between 0 a 1. We move the window one day at a time and fit the model each shift. For tuning the hyperparamters $p$, $d$, and $q$ (see (\ref{eq:arima})), we carry out an exhaustive search in $p \in \{1,2,3\}$, $d \in \{0,1,2\}$, and $q \in \{1,2,3\},$ and choose the best $(p,d,q)$ each step according to the Akaike information criterion (AIC). Likewise, the neural network models use the 10-day sliding window as input and forecast the next day data. For tuning the learning rate, we randomly choose an AWS (Biotopo), and select the one that outputs the lowest RMSE in $\lambda \in \{1,10^{-1},10^{-2},10^{-3},10^{-4}\}.$ We set $\lambda = 10^{-2}$ then. For updating the network parameters (weights and biases), we use batches of 10 consecutive windows paired with their corresponding next-day data. As loss function, we use the mean square error (MSE) that compares the next day prediction and ground-truth. We do not introduce additional regularization terms to the loss function such as dropouts. We compute the forecasting performance progressively each step before updating the networks' parameters disregarding imputed outputs.

In our experiments, we use the built-in ARIMA algorithm of the Python module ``statsmodel 0.11.1,'' and the deep-learning framework ``Pytorch 1.4.0'' for implementing the neural network models.

\section{Results and discussion}\label{S4}
This section contains the results and analyses of the application of imputation methods and forecasting models on irradiance and insolation data.  

\subsection{Imputation}\label{S4.1}
For the irradiance data analyses, our time frame of interest is from 6:00 to 18:00. Therefore, the collected data of a day is complete when the dataset contains 13 data values in this interval (one per hour). On average (for all the 12 AWS), $69,26 \%$ of training data was imputed. Table \ref{tab:imput} contains the imputation details (total number of measured and missing values on the left, and number of missing data per time series quarter on the right). For instance, in \textit{Biotopo}, there are 10.971 missing values in the first quarter (data collected between 2005/11/26 and 2008/11/04).

\begin{table}[htb!]
\centering
\caption{Details of the hourly data used in the forecasting experiments.}
\label{tab:imput}
\resizebox{0.97\textwidth}{!}{%
\begin{tabular}{@{}ll|llll|llll@{}}
\toprule
& &\multicolumn{4}{|c|}{Time series details}& \multicolumn{4}{c}{Missing values per quarter}\\ 
\toprule
AWS Name & Time frame & \# of measured & \# of missing & \% of missing & Total & $1^{sd}$quarter & $2^{nd}$quarter&$3^{th}$quarter&$4^{th}$quarter \\ 
& & values & values &  values & &\# of data &\# of data & \# of data& \# of data
\\
\midrule
Biotopo &2005/11/26-2017/08/31 &12.811 & 43.050 & $77,1 \%$ & 55.861 &10.971&10.710&12.755&8.614 \\
Altaquer & 2013/05/31-2014/08/10&2.576 & 3.105 & $54,7 \%$ & 5.681&698&922&723&762\\
Granja el Mira & 2016/07/29-2017/08/31&1.723 & 3.464 & $66,8 \%$ & 5.187&819&805&1.044&796 \\
Guapi & 2005/10/09-2017/08/30&13.247 & 43.225 & $76,5 \%$ & 56.472&10.347&10.565&13.040&9.273 \\
Cerro Páramo & 2005/11/25-2017/08/31&22.072 & 33.802 & $60,5 \%$& 55.835&9.204&7.959&8.209&8.430 \\
Viento Libre & 2005/11/13-2017/08/31&16.288 & 39.742 & $70,9 \%$ & 56.030 &12.450&9.485&8.494&9.313\\
Universidad de Nariño &2005/05/12-2017/08/31& 21.375 & 37.060 &  $63,4 \%$ & 58.435&9.467&9.655&9.300&8.638 \\
La Josefina & 2005/11/28-2017/08/31&8.815 & 47.020 & $84,2 \%$ & 55.835&12.146&13.880&11.576&9.418 \\
Botana & 2005/05/12-2017/08/31&23.036 & 35.399 & $60,6 \%$ & 58.435&8.577&9.666&8.242&8.914 \\
Paraiso & 2005/11/23-2017/08/31&17.551 & 38.349 & $68,6 \%$ & 55.900&11.259&9.540&9.530&8.020 \\
Ospina Perez& 2016/06/18-2017/08/17& 1.449 & 4.128 & $74,0 \%$ & 5.577&747&1.392&1.162&827 \\
Sandona & 2016/06/16-2017/08/17&1.454 & 4.110 & $73,9 \%$& 5.564&713&1.391&1.162&844\\
 \bottomrule
\end{tabular}%
}
\end{table}

\begin{figure}[htb!]
    \centering
    \includegraphics[width=0.8\linewidth]{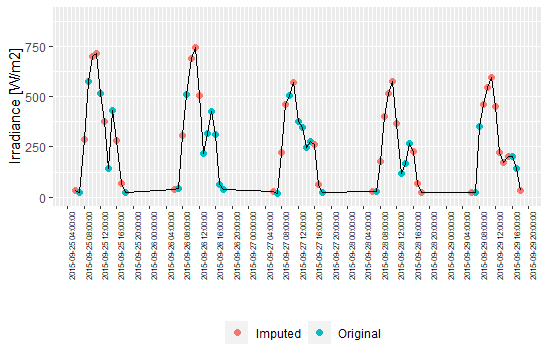}
    \caption{Irradiance data imputation \textit{La Josefina} in 2015}
    \label{fig:jf2015}
\end{figure}

\textit{La Josefina} is the AWS with most missing data. For training the forecasting models with this station, 84,2 \% data needs to be estimated. Most of its missing values are in the second quarter of the time series with 13.880 missing values $(29,5 \%)$. The longest gap with consecutive missing values is 12.999, approximately 2,7 years, and the one-size gap (one missed value between two known values) occurs 2.242 times. During 2010 there is not any register, therefore, for training the models, all values during that period were refilled. Figure \ref{fig:jf2015} shows the imputation process in a five-day time frame. As it can be seen, the imputed values follow the trend of the actual values. The statistical error shows that, in this case, the imputation process underestimates the resource by $-15,53 [Wh/m^{2}]$ (see Table \ref{tab:imperr}).    

Table \ref{tab:imperr} shows the error estimates of the imputation process in the irradiance data. According to MAE, \textit{Cerro Páramo} exhibits less extreme events. Consequently, the imputed values follow the mean trend. The RMSE shows that \textit{Sandona} and \textit{Granja el Mira} have more variability and outlier values. The average RMSE is 118,15 $[Wh/m^2]$. In the AWS located in Pacific, the RMSE is 115,84 $[Wh/m^2]$, 127,63 $[Wh/m^2]$ in the Andean zone, and 103,69 $[Wh/m^2]$ in the Amazonia zone. The MBE shows that in all AWS the resource was underestimated by $12,83 [Wh/m^2]$ on average, with a maximum in \textit{Ospina Perez} of $2,78 [Wh/m^2]$ and a minimum in \textit{Guapi} of $16,79 [Wh/m^2]$. Likewise, the MAE is 81,07 $[Wh/m^2]$ on average. In the Pacific zone, MAE is 76,75 $[Wh/m^2]$, 85,92 $[Wh/m^2]$ in the Andean zone, and 64,45 $[Wh/m^2]$ in the Amazonia zone.     

\begin{table}[htb!]
\centering
\caption{Statistical errors in the imputation process for solar irradiance forecasting}
\label{tab:imperr}
\resizebox{0.55\textwidth}{!}{%
\begin{tabular}{@{}lllll@{}}
\toprule
Regions & AWS Names & MAE $[Wh/m^2]$ & RMSE$[Wh/m^2]$ & MBE$[Wh/m^2]$ \\ \midrule
\multirow{4}{*}{Pacific} & Biotopo & 69,46 & 104,16 & -14,72 \\
 & Altaquer & 68,24 & 107,37 & -15,82 \\
 & Granja el Mira & 92,22 & 134,83 & -16,18 \\
 & Guapi & 77,09 & 117,00 & -16,79 \\ \hline
Amazonia & Cerro Páramo & 64,45 & 103,69 & -15,07 \\ \hline
\multirow{7}{*}{Andean} & Viento Libre & 80,41 & 118,04 & -11,28 \\
 & Universidad de Nariño & 85,41 & 125,66 & -9,81 \\
 & La Josefina & 83,06 & 126,03 & -15,53 \\
 & Botana & 89,84 & 131,31 & -13,83 \\
 & Paraiso & 86,46 & 123,44 & -10,2 \\
 & Ospina Perez & 89,95 & 132,07 & -2,78 \\
 & Sandona & 86,34 & 136,88 & -11,94 \\
 & Average & 81,07 & 118,15 & -12,82 \\
\bottomrule
\end{tabular}%
}
\end{table}

Table \ref{tab:inerr} shows the errors of the imputation of insolation data. The RSME shows that the Logistic model outperform Hargreaves and Samani model in the Andean and Amazonia zones, and in the Pacific zone in the AWS with lower measured data such as \textit{Altaquer} and \textit{Granja el Mira}. In the Andean zone, The average RMSE of the logistic model is $1.022,86 [Wh/m^2  day]$, and \textit{Sandona} and \textit{Viento Libre} show more variability than the other AWS located in this zone. The average MAE of the logistic model is $833,89 [Wh/m^2 day]$. In the Pacific zone, the average RMSE with logistic model is $870,38 [Wh/m^2 day]$ and with Hargreaves and Samani model is $933,50 [Wh/m^2 day]$. The average MAE of the logistic model is $686,22 [Wh/m^2 day]$ and with the Hargreaves and Samani model is $737,93 [Wh/m^2 day]$. In this zone, the logistic model outperform Hargreaves and Samani model in the AWS with lower amount of data (\textit{Altaquer} and \textit{Granja el Mira}). The MBE does not follow a patter as the RMSE and MAE; therefore, this error should be analyzed for each particular case. In the future, the time series gaps could be filled with satellite data as shows \cite{NARVAEZ}.

\begin{table}[htb!]
\centering
\caption{Statistical errors in the imputation process for insolation forecasting}
\label{tab:inerr}
\resizebox{0.65\textwidth}{!}{%
\begin{tabular}{@{}ll|ll|ll|ll@{}}
\toprule
& & \multicolumn{2}{c}{RMSE [Wh/m2 day]} & \multicolumn{2}{c}{MBE [Wh/m2 day]} &  \multicolumn{2}{c}{MAE [Wh/m2 day]} \\* \midrule
%
Regions &AWS Names &  HS & Logistic &  HS & Logistic & HS & Logistic \\
\multirow{4}{*}{Pacific} & Biotopo &  \textbf{993,64} &  1.113,48 &  \textbf{-2,01} & -37,29& \textbf{800,52} & 885,10\\
& Altaquer &761,09 & \textbf{670,59} & -333,11 &\textbf{145,14} & 596,88&\textbf{529,73} \\
& Granja el Mira &854,87 &\textbf{781,91} &-442,78 &\textbf{76,19} &684,98 & \textbf{596,66}\\
& Guapi & \textbf{878,75} &   915,53&   \textbf{-16,38} & -27,50 &  \textbf{696,35} &  733,40  \\ \hline
Amazonia & Cerro Páramo &  1.209,76 &  \textbf{1.152,72} & \textbf{21,29} &  33,37&   946,29 &   \textbf{887,34} \\ \hline
\multirow{7}{*}{Andean}& Viento Libre &  1.080,72 &  \textbf{1.077,35} &  163,13 &  \textbf{160,23} &   \textbf{861,64} &   862,86 \\
& Universidad de Nariño &   1.083,73 &   \textbf{1.019,14}&   \textbf{62,04} &  93,72 &  884,48 &  \textbf{833,30} \\
& Botana &  1.070,23 &   \textbf{1.042,68}&  \textbf{42,30} & 47,13&  881,19 &  \textbf{860,18} \\
& Josefina &   1.066,18 &  \textbf{984,75}&  \textbf{-20,58} & 22,18&  830,06 &  \textbf{760,43} \\
& Paraiso &  990,32 &  \textbf{921,32}&  -42,52 & \textbf{-14,98} &  806,64 &  \textbf{748,67} \\
& Ospina perez &987,70 &\textbf{934,93} &355,08 &\textbf{173,99} & 797,83&\textbf{736,86} \\
& Sandona &	1.375,84 &\textbf{1.179,92} &1.213,27 &\textbf{988,12} &1.249,26 &\textbf{1.034,94} \\
 \bottomrule
\multicolumn{5}{l}{Lowest values are in bold} \\ 
\end{tabular}
}
\end{table}

\subsection{Irradiance forecasting}\label{S4.2}
After applying the data imputation process, we train and test the four forecasting models: ARIMA, SL-FNN, ML-FNN, and LSTM. In total, we train 96 forecasting models (8 models for each AWS) in an hourly and daily timestamp. In order to remove the deterministic part of the time series, we use the clear sky index instead of solar irradiance or insolation data directly. To quantify the forecasting models' performance, we calculate the MAE, RMSE, and MBE error measures for each AWS.

During 2007, 2008, and 2010, rainfall increased in Colombia because of La Niña. This weather pattern was stronger in 2010 from June to December, with a peak in October and November that registered an Oceanic Niño Index (ONI) of -1,7. Likewise, in 2009, 2015 and 2016, temperature increased because of El Niño. The ONI reached an intensity of 2,6 in December 2015, being the highest value registered by the National Oceanic and Atmospheric Administration (NOAA) since 1950. These natural phenomenons affect the performance of the forecasting techniques due to changes in the cloudiness \citep{noaa}.

\begin{figure}[htb!]
    \begin{subfigure}[b]{0.97\textwidth}
    \centering
    \includegraphics[width=\textwidth]{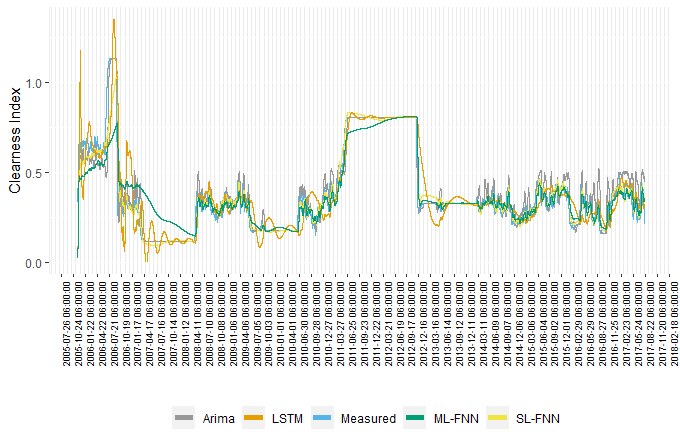}
    \caption{Complete}
    \label{fig:fbioar}
    \end{subfigure}
    \newline
    \begin{subfigure}[b]{0.49\textwidth}
    \centering
    \includegraphics[width=\textwidth]{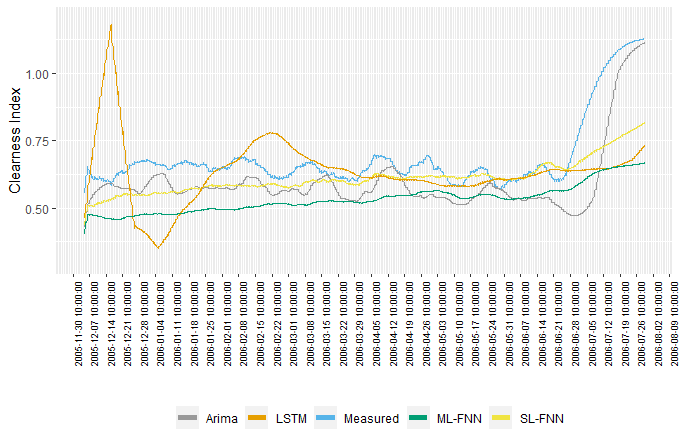}
    \caption{Starting}
    \label{fig:fbioslp}
    \end{subfigure}
    ~
    \begin{subfigure}[b]{0.49\textwidth}
    \centering
    \includegraphics[width=\textwidth]{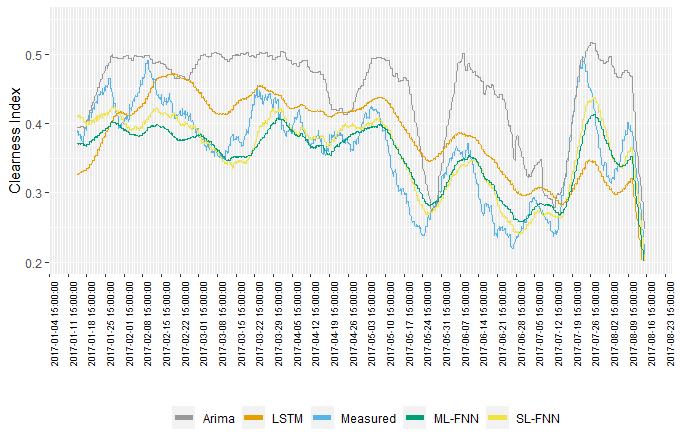}
    \caption{Ending}
    \label{fig:fbiomlp}
    \end{subfigure}
\caption{Irradiance forecasting results in \textit{Biotopo}}
\label{fig:fbio}
\end{figure}

Figure~\ref{fig:fbio} contains the hourly measured and predicted data (10-day average) in \textit{Biotopo}. Figure \ref{fig:fbio}~\textbf{(a)} shows the time series in the complete interval of acquisition. In this experiment, ARIMA, SL-FNN and ML-FNN adapt better to fast changes than LSTM, which exhibits a delayed response on these cases. To better understand the models' behavior, Figures \ref{fig:fbio} \textbf{(b)} and \textbf{(c)} show the eight-month starting and ending time intervals. ARIMA outperforms the others in the first part of the time series prediction. We attribute this observation to the large number of parameters in neural networks, which require more training data than ARIMA. On the other hand, Fig.~\ref{fig:fbio}~\textbf{(c)} shows that the neural network-based models output a more accurate prediction when observed more data.

\begin{figure}[htpb!]
    \begin{subfigure}[b]{0.97\textwidth}
    \centering
    \includegraphics[width=\textwidth]{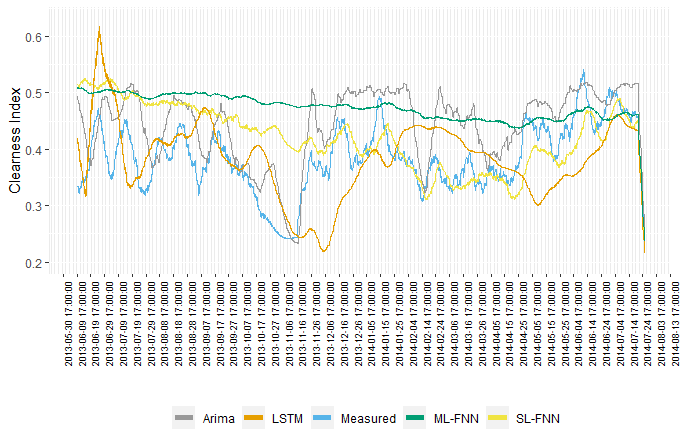}
    \caption{Complete}
    \label{fig:faltar}
    \end{subfigure}
    \newline
    \begin{subfigure}[b]{0.49\textwidth}
    \centering
    \includegraphics[width=\textwidth]{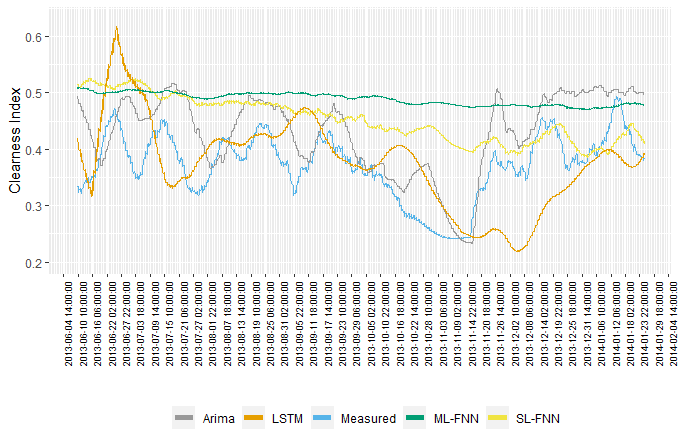}
    \caption{Starting}
    \label{fig:faltslp}
    \end{subfigure}
    ~
    \begin{subfigure}[b]{0.49\textwidth}
    \centering
    \includegraphics[width=\textwidth]{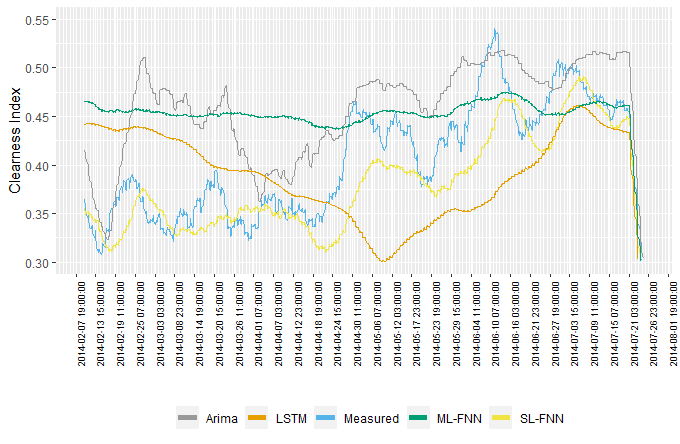}
    \caption{Ending}
    \label{fig:faltmlp}
    \end{subfigure}
\caption{Irradiance forecasting results in \textit{Altaquer}}
\label{fig:falt}
\end{figure}

Figure \ref{fig:falt} shows \textit{Altaquer} results. Figure \ref{fig:falt} \textbf{(a)} shows the time series in the complete interval of acquisition. Figure \ref{fig:falt} \textbf{(a)} shows that ML-FNN and LSTM do not have enough data inputs to adjust their weights. Figure \ref{fig:falt} \textbf{(b)} shows that ARIMA outperform the neural network-based models at the beginning; however, SL-FNN outperform ARIMA when there is six month of measures as input data on average. Figure \ref{fig:falt} \textbf{(c)} confirms that SL-FNN model outperform the others when trained with enough data.         

Table \ref{tab:h_err} contains the errors for the complete time series and errors by quarters of the time series. In \textit{Biotopo}, the MAE indicates that ARIMA is the best model, followed by SL-FNN, LSTM, and ML-FNN. We observe that ARIMA, SL-FNN, and LSTM follow a trend better than ML-FNN, and that using these models instead of ML-FNN improves the results in 25,54 \%, 21,11 \%, and 6,62 \% respectively. Analyzing the MAE by quarters, ARIMA is the best model in the first quarter (Figure \ref{fig:fbio} \textbf{(b)}). However, SL-FNN and LSTM outperform ARIMA in the next two quarters, and in the final quarter, ARIMA has the most significant error (Figure \ref{fig:fbio}~\textbf{(c)}). The RMSE shows that SL-FNN is the best model followed by LSTM, ML-FNN and ARIMA. Considering that RMSE is sensitive to outliers, these results indicate that LSTM might forecasts less atypical values. In the first quarter, ARIMA is the best model (in comparison with ML-FNN, LSTM and SL-FNN there is an error reduction of 41,73 \%, 32,43 \% and 12,68 \% respectively). Nevertheless, in the final quarter, ML-FNN, SL-FNN and LSTM outperform ARIMA by 49,35 \%, 47,42 \%, and 45,05 \%. Finally, regarding the model bias measured by MBE, LSTM is the least biased model followed by ML-FNN, SL-FNN and ARIMA. 

In \textit{Altaquer}, MAE shows that LSTM is the best model, followed by ARIMA, SL-FNN, and ML-FNN. In the first, second, and third quarters, ARIMA is the best model (maybe due to an insufficient amount of data to adjust the neural network parameters). However, in the second and third quarters, LSTM reduces the MAE difference with respect to ARIMA, and in the final quarter, LSTM outperforms ARIMA by 13,28 \%. RMSE and MAE behave similarly for this AWS. ARIMA is the best model in the first three quarters outperforming LSTM by 14,92 \%, 5,8 \% and 0,92 \%, SL-FNN by 13,63 \%, 46,45 \% and 38,62 \%, and ML-FNN by 21,87 \%, 39,62 \%, 30,03 \%. LSTM and SL-FNN outperform ARIMA in the final quarter by 10,83 \%, and 8,87 \% respectively. As a result, LSTM is the best model in the complete time series, followed by ARIMA, SL-FNN and ML-FNN. MBE shows that LSTM is the model with less bias, outperforming SL-FNN, ARIMA, and ML-FNN by 40,27 \%, 64,11 \% and 98,07 \%, respectively. 

In \textit{Granja el Mira}, MAE and RSME show that ARIMA is the best model. However, ARIMA is also the most biased model. Analyzing the models by quarters, ARIMA exhibits the best MAE and RMSE in the first three quarters. In the final quarter, SL-FNN is the best model. Regarding the MBE, ARIMA exhibits less bias in the first three quarters, and SL-FNN in the final quarter. However, the MBE in the complete time series shows ML-FNN as the least biased model. MBE is a measure that in long-term could compensate overestimation and underestimation periods, as shown in Table \ref{tab:estaerr4}.     

In \textit{Guapi}, SL-FNN exhibits the lowest error regarding MAE and RMSE followed by ARIMA, ML-FNN, and LSTM. In the first and fourth quarters, SL-FNN shows the lowest MAE and RMSE error. ARIMA is the best option in the second and third quarters. Regarding the bias, SL-FNN shows the lowest value, followed by LSTM, ML-FNN, and ARIMA.  

In \textit{Viento Libre}, the AWS located in the Andean zone with less cloudiness (see Figure \ref{fig:clinaws}), SL-FNN has the lowest MAE and RMSE, followed by ARIMA, ML-FNN, and LSTM. In the first three-quarters, ARIMA is the best model, and in the final quarter, this model has the highest error. In the first quarter, ARIMA and SL-FNN have similar RMSE values, and in the second and third quarters, ARIMA outperforms the others. Nevertheless, in the final quarter, the neural network-based models outperform ARIMA. ARIMA is less biased during the first three quarters. However, LSTM outperforms ARIMA by 99,49 \% in the last quarter. 

In \textit{Universidad de Nariño}, MAE shows that the neural network-based models outperform ARIMA. However, in the analysis by quarters, ARIMA outperforms neural-network-based models during the first three quarters. Regarding the RMSE, the neural network-based models outperform ARIMA. Nevertheless, the difference between ARIMA and LSTM is approximately 0,81 \%. In the first three quarters, ARIMA has a lower error than the neural network-based models. The MBE shows LSTM as the least biased model. Besides, LSTM has the lowest bias in the first and fourth quarters. In \textit{Botana}, the neural network-based models exhibit lower MAE and RMSE than ARIMA. The MBE shows that ML-FNN is the most biased model, and LSTM, SL-FNN, and ARIMA reduce the bias by 96,83 \%, 77,55 \%, and 77,25 \%, respectively. MAE shows that ARIMA is the best model in the first three quarters, and SL-FNN is the best in the fourth quarter. ARIMA, SL-FNN, and ML-FNN have a similar RMSE with a difference of 0,06 \% on average in the first quarter. In the following two quarters, ARIMA is the best model. SL-FNN is the best model in the fourth quarter. MBE shows that ML-FNN is more biased than LSTM, SL-FNN and ARIMA by 96,83 \%, 77,55 \% and 77,25 \% respectively. \textit{Universidad de Nariño} and \textit{Botana} are located close to each other, as illustrated in Figure \ref{fig:aws3}, and the altitude difference is 194 MASL. However, \textit{Botana} presents more cloudiness (Fig.~\ref{fig:clinaws}). Therefore, this condition could affect the performance of the forecasting models. In \textit{Universidad de Nariño} the MPE is 23,53 \%, 19,52 \%, 17,77 \% and 12,84 \% for ARIMA, ML-FNN, LSTM and SL-FNN respectively. In \textit{Botana}, MPE is 41,88 \%, 29,61 \%, 26,34 \%, and 23,21 \% for ARIMA, ML-FNN, LSTM and SL-FNN respectively. These results show that cloudy conditions increase the statistical errors. ARIMA present the largest increase with 18,35 \%, followed by SL-FNN (10,37 \%), ML-FNN (10,09 \%), and LSTM (8,57 \%).

\textit{La Josefina} is a critical case regarding missing data (almost 80 \% of the time series). MAE shows that SL-FNN better fits the mean trend of the measured data, followed by LSTM, ARIMA, and ML-FNN. ARIMA is the model with the highest RMSE. By using SL-FNN, LSTM and ML-FNN, RMSE reduces by 90,19 \%, 88,29 \% and 87,82 \% respectively. MBE shows that LSTM is less biased model. In the first quarter, ARIMA is the best model regarding MAE, and SL-FNN is the best model in the second, third, and fourth quarters. The RMSE shows that SL-FNN has the lowest error in all four quarters.

In \textit{Para\'iso}, SL-FNN has the lowest MAE and RMSE values in the complete time series analysis. ARIMA exhibits the lowest MAE and RMSE error in the first three quarters. In the last quarter, SL-FNN is the best model, followed by ML-FNN, LSTM, and ARIMA. Accordint to MBE, LSTM has the lowest bias, followed by SL-FNN, ML-FNN, and ARIMA.

In \textit{Ospina Perez}, MAE and RSME show that ARIMA and ML-FNN are the best and worst model, respectively. In the first quarter, SL-FNN has the lowest MAE. From the second to the fourth quarter, LSTM has the lowest MAE. The RMSE indicates that ARIMA is the best model in the first quarter and LSTM is the best model from the second to the fourth quarter. Additionally, LSTM is the less biased model followed by SL-FNN, ARIMA, and ML-FNN in the complete time series. In \textit{Sandon\'a}, MAE and RMSE show that ARIMA is the best model. The MBE indicates that LSTM is the less biased model, followed by ARIMA. 

In \textit{Cerro Páramo}, MAE and RSME show that SL-FNN is the best model, followed by ML-FNN, LSTM, and ARIMA. The neural network-based models outperform ARIMA in the first quarter. This result is different to the ones obtained with the AWS analyzed above, where ARIMA outperforms the neural network-based models during the first three quarters. The cloud cover amount of \textit{Cerro P\'aramo} could explain these results. SL-FNN is the least biased model, followed by LSTM, ML-FNN, and ARIMA.

SL-FNN outperforms the other forecasting models (regarding RMSE and MAE) for the AWS with more than one year of measures in the Andean zone (\textit{La Josefina}, \textit{Viento Libre}, \textit{Universidad de Nari\~no}, \textit{Botana}, and \textit{El Para\'iso}). On the other hand, ARIMA exhibits the best performance in the AWS with approximately one year of data (\textit{Sandon\'a} and \textit{\'Ospina P\'erez}), and LSTM is the least biased model for all the AWS in the Andean Zone. SL-FNN also outperforms the others in \textit{Cerro Páramo} (Amazonia), which is  the AWS with more cloudy days (see Fig.~\ref{fig:clinaws}). LSTM outperforms the others in \textit{Altaquer} (Pacific). SL-FNN is the best model in \textit{Guapi} (Pacific). In the Pacific, in \textit{Granja el Mira} and \textit{Biotopo} there is not a model that outperforms the others for all the error estimates. However, the experiments show that ARIMA is the best model for \textit{Granja el Mira} and that SL-FNN exhibits the best performance in \textit{Biotopo} regarding RMSE and MAE.  

\begin{landscape}
\fontsize{7}{11}\selectfont{
\begin{longtable}[c]{p{0.8cm}l|lll|llll|llll|llll}
\hline
 &  & \multicolumn{3}{|c|}{Complete} & \multicolumn{4}{c|}{MAE} & \multicolumn{4}{c|}{RMSE} & \multicolumn{4}{c}{MBE} \\ \cline{3-17}
\endhead
AWS & Model & MAE & MBE & RMSE & Q1 & Q2 & Q3 & Q4 & Q1 & Q2 & Q3 & Q4 & Q1 & Q2 & Q3 & Q4 \\ \hline
\parbox[t]{2mm}{\multirow{4}{*}{\rotatebox[origin=c]{90}{Biotopo}}}  & ARIMA & \textbf{0.093} & 0.014 & 0.193 & \textbf{0.392} & 0.566 & 0.829 & 0.151 & \textbf{0.325} & 0.411 & 0.496 & 0.306 & -0.025 & -0.558 & -0.812 & 0.044 \\ \cline{2-17} 
 & SL-FNN & 0.098 & -0.003 & \textbf{0.153} & 0.523 & 0.527 & 0.828 & 0.119 & 0.396 & 0.402 & 0.522 & 0.161 & -0.105 & -0.509 & -0.793 & 0.002 \\ \cline{2-17} 
 & LSTM & 0.116 & \textbf{0.002} & 0.179 & 0.659 & \textbf{0.520} & \textbf{0.746} & 0.129 & 0.506 & \textbf{0.393} & \textbf{0.473} & 0.168 & \textbf{0.016} & \textbf{-0.494} & \textbf{-0.695} & \textbf{0.001} \\ \cline{2-17} 
 & ML-FNN & 0.125 & -0.002 & 0.184 & 0.858 & 0.944 & 1.351 & \textbf{0.116} & 0.558 & 0.693 & 0.820 & \textbf{0.155} & 0.041 & -0.898 & -1.259 & -0.001 \\ \hline
\parbox[t]{2mm}{\multirow{4}{*}{\rotatebox[origin=c]{90}{Altaquer}}} & ARIMA & 0.182 & 0.058 & 0.228 & \textbf{0.733} & \textbf{0.338} & \textbf{0.590} & 0.199 & \textbf{0.449} & \textbf{0.284} & \textbf{0.392} & 0.240 & 0.236 & \textbf{0.072} & \textbf{0.057} & 0.042 \\ \cline{2-17} 
 & SL-FNN & 0.186 & 0.037 & 0.232 & 0.860 & 0.684 & 0.978 & 0.177 & 0.527 & 0.530 & 0.639 & 0.219 & 0.413 & 0.557 & 0.723 & -0.033 \\ \cline{2-17} 
 & LSTM & \textbf{0.175} & \textbf{-0.002} & \textbf{0.223} & 0.844 & 0.411 & 0.624 & \textbf{0.172 }& 0.534 & 0.314 & 0.398 & \textbf{0.214} & \textbf{0.175} & 0.265 & 0.330 & -0.063 \\ \cline{2-17} 
 & ML-FNN & 0.216 & 0.088 & 0.259 & 0.978 & 0.670 & 0.957 & 0.200 & 0.574 & 0.493 & 0.584 & 0.241 & 0.433 & 0.383 & 0.384 & \textbf{0.016} \\ \hline
\parbox[t]{2mm}{\multirow{4}{*}{\rotatebox[origin=c]{90}{\begin{tabular}[c]{@{}l@{}}Granja el \\ Mira\end{tabular}}}}  & ARIMA & \textbf{0.169} & -0.013 & \textbf{0.219} & \textbf{0.603} & \textbf{0.135} & \textbf{0.177} & 0.184 & \textbf{0.412} & \textbf{0.107} & \textbf{0.119} & 0.227 & \textbf{-0.048} & \textbf{-0.126} & \textbf{-0.156} & -0.029 \\ \cline{2-17} 
 & SL-FNN & 0.177 & 0.011 & 0.228 & 0.924 & 0.926 & 1.267 & \textbf{0.153} & 0.572 & 0.699 & 0.797 & \textbf{0.194} & 0.205 & 0.922 & 1.258 & \textbf{-0.010} \\ \cline{2-17} 
 & LSTM & 0.268 & -0.007 & 0.351 & 1.343 & 0.686 & 1.027 & 0.265 & 0.858 & 0.495 & 0.608 & 0.371 & -0.277 & 0.672 & 0.998 & 0.082 \\ \cline{2-17} 
 & ML-FNN & 0.217 & \textbf{0.003} & 0.272 & 1.033 & 0.442 & 0.790 & 0.189 & 0.638 & 0.397 & 0.555 & 0.234 & 0.181 & 0.439 & 0.786 & -0.049 \\ \hline
\parbox[t]{2mm}{\multirow{4}{*}{\rotatebox[origin=c]{90}{Guapi}}} & ARIMA & 0.130 & 0.010 & 0.205 & \textbf{0.490} & \textbf{0.255} & \textbf{0.354} & 0.177 & 0.357 & \textbf{0.205} & \textbf{0.244} & 0.251 & 0.088 & \textbf{0.162} & \textbf{0.168} & 0.014 \\ \cline{2-17} 
 & SL-FNN & \textbf{0.106} & \textbf{0.000} & \textbf{0.157} & 0.479 & 0.488 & 0.647 & \textbf{0.140} & \textbf{0.331} & 0.386 & 0.434 & \textbf{0.193} & \textbf{-0.004} & -0.393 & -0.456 & -0.004 \\ \cline{2-17} 
 & LSTM & 0.197 & 0.002 & 0.336 & 0.550 & 0.320 & 0.581 & 0.273 & 0.386 & 0.280 & 0.392 & 0.379 & 0.016 & -0.265 & -0.471 & 0.072 \\ \cline{2-17} 
 & ML-FNN & 0.152 & -0.003 & 0.215 & 0.752 & 0.413 & 0.731 & 0.156 & 0.480 & 0.399 & 0.556 & 0.209 & 0.131 & -0.193 & -0.482 & \textbf{0.003} \\ \hline
\parbox[t]{2mm}{\multirow{4}{*}{\rotatebox[origin=c]{90}{\begin{tabular}[c]{@{}l@{}}Viento \\ Libre \end{tabular}}}} & ARIMA & 0.145 & -0.014 & 0.206 & \textbf{0.182} & \textbf{0.147} & \textbf{0.231} & 0.176 & \textbf{0.205} & \textbf{0.149} & \textbf{0.201} & 0.226 & \textbf{0.004} & \textbf{0.093} & \textbf{0.125} & -0.024 \\ \cline{2-17} 
 & SL-FNN & \textbf{0.117} & -0.004 & \textbf{0.164} & 0.273 & 0.284 & 0.485 & \textbf{0.124} & 0.243 & 0.252 & 0.347 & \textbf{0.164} & -0.026 & -0.183 & -0.283 & -0.002 \\ \cline{2-17} 
 & LSTM & 0.156 & \textbf{0.000} & 0.247 & 0.513 & 0.424 & 0.633 & 0.152 & 0.652 & 0.329 & 0.412 & 0.196 & 0.041 & -0.344 & -0.474 & \textbf{0.000} \\ \cline{2-17} 
 & ML-FNN & 0.153 & -0.007 & 0.203 & 0.647 & 0.758 & 0.976 & 0.138 & 0.465 & 0.600 & 0.655 & 0.176 & -0.126 & -0.614 & -0.689 & -0.008 \\ \hline
\parbox[t]{2mm}{\multirow{4}{*}{ \rotatebox[origin=c]{90}{\begin{tabular}[c]{@{}l@{}}Universidad \\ de Nariño\end{tabular}}}} & ARIMA & 0.173 & -0.015 & 0.228 & \textbf{0.484} & \textbf{0.125} & \textbf{0.199} & 0.195 & \textbf{0.319} & \textbf{0.117} & \textbf{0.158} & 0.241 & -0.035 & \textbf{-0.084} & \textbf{-0.116} & -0.014 \\ \cline{2-17} 
 & SL-FNN & \textbf{0.144} & -0.008 & \textbf{0.194} & 0.511 & 0.613 & 0.977 & 0.136 & 0.334 & 0.469 & 0.614 & \textbf{0.178} & -0.045 & -0.591 & -0.934 & -0.004 \\ \cline{2-17} 
 & LSTM & 0.169 & \textbf{0.000} & 0.226 & 0.560 & 0.495 & 0.758 & 0.147 & 0.389 & 0.359 & 0.452 & 0.188 & \textbf{-0.022} & -0.484 & -0.735 & \textbf{0.000} \\ \cline{2-17} 
 & ML-FNN & 0.158 & 0.000 & 0.209 & 0.519 & 0.311 & 0.521 & \textbf{0.141} & 0.343 & 0.280 & 0.382 & 0.183 & -0.126 & -0.220 & -0.339 & -0.006 \\ \hline
 \parbox[t]{2mm}{\multirow{4}{*}{\rotatebox[origin=c]{90}{Josefina}}}  & ARIMA & 0.091 & -0.016 & 1.294 & \textbf{0.299} & 0.160 & 0.218 & 0.163 & 0.293 & 0.134 & 0.159 & 0.218 & 0.007 & -0.138 & -0.173 & -0.019 \\ \cline{2-17} 
 & SL-FNN & \textbf{0.074} & 0.003 & \textbf{0.127} & 0.365 & \textbf{0.083} & \textbf{0.152} & \textbf{0.124} & \textbf{0.286} & \textbf{0.099} & \textbf{0.140} & \textbf{0.168} & 0.048 & \textbf{-0.006} & \textbf{-0.026} & \textbf{0.002} \\ \cline{2-17} 
 & LSTM & 0.089 & \textbf{0.000} & 0.151 & 0.370 & 0.259 & 0.337 & 0.156 & 0.302 & 0.210 & 0.237 & 0.205 & \textbf{-0.005} & -0.181 & -0.181 & -0.003 \\ \cline{2-17} 
 & ML-FNN & 0.103 & 0.018 & 0.158 & 0.640 & 0.274 & 0.441 & 0.134 & 0.412 & 0.220 & 0.292 & 0.174 & 0.239 & -0.008 & 0.091 & -0.004 \\ \hline
\parbox[t]{2mm}{\multirow{4}{*}{\rotatebox[origin=c]{90}{Botana}}} & ARIMA & 0.178 & -0.003 & 0.228 & \textbf{0.681} & \textbf{0.125} & \textbf{0.210} & 0.183 & 0.433 & \textbf{0.112} & \textbf{0.154} & 0.234 & \textbf{0.000} & \textbf{-0.015} & \textbf{0.009} & -0.007 \\ \cline{2-17} 
 & SL-FNN & \textbf{0.149} & -0.003 & \textbf{0.194} & 0.682 & 0.455 & 0.707 & \textbf{0.136} & 0.433 & 0.344 & 0.443 & \textbf{0.178} & -0.017 & -0.032 & -0.268 & -0.002 \\ \cline{2-17} 
 & LSTM & 0.162 & \textbf{0.000} & 0.211 & 0.715 & 0.268 & 0.461 & 0.153 & 0.479 & 0.226 & 0.311 & 0.194 & -0.009 & -0.254 & -0.434 & \textbf{0.000} \\ \cline{2-17} 
 & ML-FNN & 0.153 & 0.012 & 0.196 & 0.684 & 0.179 & 0.251 & 0.142 & \textbf{0.433} & 0.150 & 0.183 & 0.181 & 0.160 & -0.141 & -0.173 & -0.001 \\ \hline
\parbox[t]{2mm}{\multirow{4}{*}{\rotatebox[origin=c]{90}{Paraiso}}} & ARIMA & 0.156 & -0.022 & 1.578 & \textbf{0.286} & \textbf{0.114} & \textbf{0.207} & 0.189 & \textbf{0.245} & \textbf{0.106} & \textbf{0.148} & 0.238 & -0.037 & \textbf{0.014} & \textbf{0.050} & -0.010 \\ \cline{2-17} 
 & SL-FNN & \textbf{0.123} & -0.004 & \textbf{0.172} & 0.318 & 0.375 & 0.534 & \textbf{0.141} & 0.254 & 0.282 & 0.335 & \textbf{0.184} & -0.030 & -0.344 & -0.472 & \textbf{-0.003} \\ \cline{2-17} 
 & LSTM & 0.138 & \textbf{-0.001} & 0.192 & 0.329 & 0.426 & 0.626 & 0.167 & 0.261 & 0.311 & 0.377 & 0.216 & \textbf{-0.014} & -0.400 & -0.574 & -0.004 \\ \cline{2-17} 
 & ML-FNN & 0.137 & -0.009 & 0.186 & 0.419 & 0.317 & 0.550 & 0.148 & 0.321 & 0.302 & 0.418 & 0.190 & -0.176 & -0.116 & -0.317 & -0.008 \\ \hline
\parbox[t]{2mm}{\multirow{4}{*}{ \rotatebox[origin=c]{90}{\begin{tabular}[c]{@{}l@{}}Ospina \\ Perez \end{tabular}}}} & ARIMA & \textbf{0.123} & -0.033 & \textbf{0.193} & \textbf{0.841} & 0.353 & 0.530 & 0.180 & \textbf{0.519} & 0.261 & 0.325 & 0.240 & -0.223 & -0.162 & -0.148 & -0.045 \\ \cline{2-17} 
 & SL-FNN & 0.163 & -0.026 & 0.219 & 0.839 & 0.288 & 0.504 & 0.171 & 0.524 & 0.294 & 0.411 & 0.225 & -0.317 & 0.046 & 0.163 & 0.062 \\ \cline{2-17} 
 & LSTM & 0.169 & \textbf{-0.015} & 0.236 & 0.924 & \textbf{0.200} & \textbf{0.381} & \textbf{0.166} & 0.588 & \textbf{0.207} & \textbf{0.292} & \textbf{0.217} & \textbf{-0.022} & \textbf{0.002} & \textbf{0.021} & -0.070 \\ \cline{2-17} 
 & ML-FNN & 0.213 & -0.129 & 0.267 & 0.842 & 0.418 & 0.585 & 0.175 & 0.536 & 0.315 & 0.367 & 0.225 & -0.441 & -0.314 & -0.379 & \textbf{-0.041} \\ \hline
\parbox[t]{2mm}{\multirow{4}{*}{\rotatebox[origin=c]{90}{Sandona}}} & ARIMA & \textbf{0.132} & -0.015 & \textbf{0.199} & \textbf{0.817} & \textbf{0.303} & \textbf{0.536} & \textbf{0.158} & \textbf{0.503} & \textbf{0.275} & \textbf{0.382} & \textbf{0.205} & -0.128 & -0.088 & -0.108 & \textbf{0.005} \\ \cline{2-17} 
 & SL-FNN & 0.284 & -0.159 & 0.394 & 0.875 & 0.317 & 0.595 & 0.162 & 0.545 & 0.312 & 0.443 & 0.210 & -0.168 & 0.028 & 0.098 & -0.027 \\ \cline{2-17} 
 & LSTM & 0.259 & \textbf{-0.013} & 0.328 & 1.124 & 0.308 & 0.572 & 0.213 & 0.711 & 0.295 & 0.414 & 0.255 & \textbf{-0.006} & \textbf{0.011} & 0.069 & 0.064 \\ \cline{2-17} 
 & ML-FNN & 0.328 & -0.150 & 0.410 & 0.894 & 0.619 & 0.879 & 0.198 & 0.556 & 0.484 & 0.581 & 0.244 & -0.087 & 0.434 & \textbf{0.507} & 0.062 \\ \hline
\parbox[t]{2mm}{\multirow{4}{*}{ \rotatebox[origin=c]{90}{\begin{tabular}[c]{@{}l@{}}Cerro \\ Páramo\end{tabular}}}} & ARIMA & 0.210 & 0.074 & 0.349 & 0.822 & \textbf{0.129} & \textbf{0.188} & 0.213 & 1.026 & \textbf{0.103} & \textbf{0.128} & 0.296 & 0.254 & \textbf{-0.111} & \textbf{-0.151} & 0.069 \\ \cline{2-17} 
 & SL-FNN & \textbf{0.136} & \textbf{0.001} & \textbf{0.186} & \textbf{0.614} & 0.494 & 0.774 & \textbf{0.131} & \textbf{0.409} & 0.388 & 0.506 & \textbf{0.185} & \textbf{0.009} & -0.485 & -0.758 & -0.005 \\ \cline{2-17} 
 & LSTM & 0.177 & 0.002 & 0.277 & 0.677 & 0.362 & 0.560 & 0.260 & 0.454 & 0.266 & 0.338 & 0.430 & 0.018 & -0.349 & -0.536 & \textbf{0.001} \\ \cline{2-17} 
 & ML-FNN & 0.156 & 0.003 & 0.214 & 0.756 & 0.512 & 0.801 & 0.162 & 0.469 & 0.391 & 0.507 & 0.249 & 0.149 & -0.496 & -0.770 & -0.024 \\ \hline
\caption{Irradiance statistical errors of the one day-ahead forecasting process}
\label{tab:h_err}\\
\end{longtable}
}
\end{landscape}

\subsection{Insolation forecasting}\label{S4.3}
In this study,  we also apply the forecasting models for the one-ahead day prediction of daily solar insolation. The daily solar insolation is the sum of hourly irradiances measured between 6:00  and 18:00 hours. Table \ref{tab:d_err} shows the errors for a daily time stamp.

In \textit{Biotopo}, MAE shows that LSTM is the best option for daily global solar irradiance forecasting, followed by ML-FNN, ARIMA, and SL-FNN. It indicates that LSTM fits the time series' mean trend better than ML-FNN, ARIMA, and SL-FNN. As a result, using LSTM reduces the MAE error by 6,41 \%, 17,73 \%, and 20,23 \% in comparison with ML-FNN, ARIMA and SL-FNN. Analyzing the MAE by quarters, LSTM outperforms the other models in the first and fourth quarters, and ARIMA is the best model in the second and third quarters. The RMSE shows that LSTM is the model with the lowest variability, followed by ML-FNN, ARIMA, and SL-FNN. When using LSTM, ML-FNN or ARIMA instead of SL-FNN, the RMSE decreases by 23,33 \%, 16,41 \% and 9,07 \%, correspondingly. Furthermore, LSTM presents less bias than the other models and reduces the bias by 89,98 \%, 44,88 \%, and 33,47 \% in comparison with SL-FNN, ARIMA, and ML-FNN.  

In \textit{Altaquer}, with one year of measurements on average, LSTM outperforms the other models. MAE shows that LSTM and ARIMA are the best options. LSTM improves the results by 58,43 \%, 48,43 \% and 6,75 \% in comparison with SL-FNN, ML-FNN and ARIMA. Also, the analysis by quarters shows that LSTM has the best behavior. RSME shows LSTM as the best option, followed by ARIMA, ML-FNN, and SL-FNN. LSTM reduces the RMSE in 60,01 \%, 53,85 \%, and 8,18 \% in comparison with SL-FNN, ML-FNN and ARIMA, respectively. LSTM is the less biased model followed by ARIMA, ML-FNN, and SL-FNN. MAPE shows that LSTM has an error of 32,63 \%, ARIMA of 38,47 \%, ML-FNN of 65,95 \% and SL-FNN of 72,53 \%. The large proportion of missing data and short length of the time series might explain the large errors obtained in this case.

LSTM is the best model in \textit{Granja el Mira} considering MAE, RMSE, and MBE. MAE shows that LSTM improves the performance by 27,71 \%, 13,55 \%, and 11,60 \% in comparison with SL-FNN, ARIMA and ML-FNN. RMSE shows that LSTM has less variability than ML-FNN, ARIMA, and SL-FNN by 12 \%, 17,75 \% and 30,61 \%. The analysis by quarters shows that MAE and RMSE have similar results. LSTM is the less biased model followed by ARIMA, ML-FNN, and SL-FNN. MAPE shows that LSTM has an error of 26,16 \%, ML-FNN of 29,10 \%, ARIMA of 30,75 \% and SL-FNN of 35,71 \%. 

In \textit{Guapi}, the MAE, RSME, and MBE show that the neural network-based models outperform ARIMA. Contrasting these models with the ARIMA model, the MAE error is reduced by 19,73 \%, 15,07 \% and 14,12 \% with LSTM, SL-FNN and ML-FNN respectively. RMSE shows that LSTM is the best model, followed by SL-FNN, ML-FNN, and ARIMA. LSTM reduces the RMSE value by 1,15 \%, 2,09 \% and 20,85 \% in comparison with SL-FNN, ML-FNN and ARIMA respectively. The MBE results shows that LSTM is the less biased model, followed by SL-FNN, ML-FNN, and ARIMA. Therefore, the statistical errors show that the neural network-based models outperform ARIMA.

In \textit{Cerro Páramo} the MAE, RMSE and MBE shows SL-FNN as the best option for daily solar insolation. Using SL-FNN instead ML-FNN, ARIMA and LSTM improve MAE in 5,84 \%, 10,55 \% and 10,82 \%. In the analysis by quarters, SL-FNN is the best option, followed by ARIMA in the three first quarters and by ML-FNN in the last quarter. RSME shows that the SL-FNN model reduces the variability 6,26 \%, 9,96 \%, and 16,69 \% in comparison with ML-FNN, LSTM, and ARIMA models respectively. MBE result shows that ARIMA has more bias than ML-FNN, LSTM, and SL-FNN, increasing the error on average by 65,56 \%, 88,80 \%, and 92,50 \%, respectively. MAPE shows that SL-FNN is the best model with an error of 38,03 \% followed by ML-FNN with 39,52 \%, ARIMA with 40,90 \% and LSTM with 42,75 \%. 

In \textit{Viento Libre}, the statistical errors show ML-FNN as the model with less MAE, RMSE, and MBE errors, followed by LSTM, SL-FNN, and ARIMA. MAE shows that ML-FNN and LSTM have a similar error with a difference of 0,16 \%. Considering RMSE, ARIMA model exhibits the largest errors, and using ML-FNN, LSTM or SL-FNN reduce the error by 19,13 \%, 18,94 \% or 2,52 \% respectively. MBE shows that the models with less bias are ML-FNN and LSTM. These models reduce the bias by 77,95 \% and 99,11 \% compared to SL-FNN and ARIMA. Therefore, ML-FNN and LSTM reduce the bias in comparison with SL-FNN and ARIMA. MAPE shows that on average LSTM has an error of 19,11 \%, ML-FNN of 19,25 \%, SL-FNN of 21,95 \% and ARIMA of 23,11 \%. 

In \textit{Universidad de Nariño}, ML-FNN and SL-FNN outperform the other models. MAE shows that ML-FNN has the lowest error in both, the complete time series, and quarters analysis, followed by SL-FNN. RSME shows that using ML-FNN instead of SL-FNN, ARIMA and LSTM reduces the outliers by 1,9 \%, 18,11 \%, and 25,4 \%, respectively. MBE presents the neural network-based models as the models with less bias. MAPE shows that ARIMA has an error of 28,21 \%, LSTM of 24,89 \%, SL-FNN of 23,64 \% and ML-FNN of 23,49 \%.   

In \textit{Botana}, the MAE and RMSE show that ML-FNN is the best model. ML-FNN reduces the MAE error, in comparison with SL-FNN, LSTM and ARIMA, in 2,02 \%, 3,54 \% and 19,28 \% respectively. RMSE shows that ML-FNN has less variability than SL-FNN, LSTM and ARIMA reduce the error by 1,70 \%, 6,78 \% and 19,80 \% respectively. In \textit{Botana}, LSTM is less biased than SL-FNN, ML-FNN and ARIMA. MAPE shows that on average ML-FNN has an error of 27,23 \%, SL-FNN of 27,65 \%, LSTM of 28,21 \% and ARIMA of 33,52 \%. The neural network-based models outperform the ARIMA model. Furthermore, in \textit{Universidad de Nariño} and \textit{Botana}, which are AWS located close to each other, the MAE and MBE have similar values with the four forecasting models. However, RMSE shows that in cloudy environments, the neural network-based models' variability is lower than ARIMA. Additionally, MAPE shows that the neural network-based models have an error lower than 20 \% on average. 

In \textit{La Josefina}, the MAE, RMSE, and MBE show that LSTM is the best models followed by ML-FNN, ARIMA, and SL-FNN. Using LSTM instead ML-FNN, ARIMA and SL-FNN reduce the MAE in 5,54 \%, 17,62 \% and 23,34 \% respectively. Considering RMSE, LSTM, ML-FNN and ARIMA improve the results by 33,42 \%, 31,43 \%, and 17,30 \% in comparison with SL-FNN. LSTM has the lower bias, followed by ML-FNN, ARIMA, and SL-FNN. MAPE shows that LSTM with an error of 14,39 \%, ML-FNN of 15,42 \%, ARIMA of 17,52 \% and SL-FNN of 19,06 \%. MAPE values are lower than 20 \%; however, the amount of missing and imputed data of this AWS is considerable, which could affect the error measurements.   

In \textit{Paraiso}, the neural network-based models outperform ARIMA. MAE shows that LSTM is the best model, followed by ML-FNN, SL-FNN, and ARIMA. Also, LSTM reduces the MAE by 3,20 \%, 4 \%, and 19,59 \% in comparison with ML-FNN, SL-FNN and ARIMA. Considering RMSE, the LSTM model is the best, followed by ML-FNN, SL-FNN, and ARIMA. Also, LSTM improves the result by 2,22 \% SL-FNN 4,63 \% and 19,88 \% in comparison with ML-FNN, SL-FNN and ARIMA models. Additionally, LSTM is the model with the lowest bias, followed by SL-FNN, ML-FNN, and ARIMA. Also, the neural network-based model reduces the bias on average by 82,52 \% in comparison with ARIMA. MAPE shows that on average the error of LSTM is 19,84 \%, ML-FNN of 20,71 \%, SL-FNN of 20,86 \& and ARIMA of 24,63 \%. In this case, LSTM is the only model with average error below 20 \%.   

In \textit{Ospina Perez}, ML-FNN has the largest error due to the low amount of training data. Also, this AWS has a high amount of imputed values resulting in a low amount of data for statistical errors measures calculation. Considering the MAE and RMSE, SL-FNN is the best forecasting model, followed by ARIMA and LSTM. The improvement obtained with using SL-FNN, ARIMA, or LSMT instead of ML-FNN model is 59,53 \% in MAE and 55,28 \% in RSME on average. Regarding the MBE, the model with less bias is SL-FNN followed by LSTM. SL-FNN reduces the bias by 2,66 \%, 8,83 \% and 98,85 \% in comparison with LSTM, ARIMA and ML-FNN. The MAPE shows that SL-FNN has the lowest error with 13,95 \%, followed by ARIMA with 16,30 \%, LSTM with 20,92 \% and ML-FNN with 43,14 \%. 

In \textit{Sandona}, ML-FNN has the largest MAE, RMSE, and MBE values. This AWS has one year of measurements on average, as \textit{Ospina Perez}. Therefore, in the Andean zone, ML-FNN needs as input more than one year of measurements to describe the variability of this zone's global solar insolation. MAE shows that LSTM is the best option, followed by SL-FNN, ARIMA, and ML-FNN. The RMSE presents the LSTM model as the best forecasting option reducing the error by 4,61 \% in comparison with SL-FNN, 12,47 \% in comparison with ARIMA, and 33,02 \% in comparison with ML-FNN. Regarding the bias, the SL-FNN and LSTM have the lowest bias. MAPE shows that LSTM has an average error of 19,38 \%, SL-FNN of 21,52 \%, ARIMA of 23,58 \% and ML-FNN of 30,07 \%.

The AWS used to compare the forecasting models are located in an Intertropical Convergence Zone, which implies there is high cloudiness amount and precipitation. This situation could affect the models' accuracy because it increases randomness. Additionally, the significant amount of missing data might also affect the forecasting accuracy.  

In the Pacific zone,  \textit{Biotopo} and \textit{Altaquer} (located in the Pacific foothill) face high humidity and rainy environments. Consequently, MPE and MAPE show that the models have a large error for both, irradiance and insolation forecasting. In this case, LSTM is the only model that has an error close to 30 \%. In the other AWS of the Pacific zone, SL-FNN is the best option in the irradiance forecasting when there are more than one year of measurements. LSTM is the best option for all the AWS in a insolation forecasting. As future work, changes in the memory of LSTM could be introduced to analyze the capability of this to follow the clear sky index variability of the global solar irradiance. 

In the Andean zone, SL-FNN is the best model considering MAE, RMSE, and MPE in five of the seven AWS in the irradiance forecasting. However, LSTM is the less biased model in all cases. In the insolation forecasting, the neural network-based models outperform ARIMA. In \textit{Cerro Páramo} that is located in the Amazonia zone and is the most cloudy AWS, SL-FNN is the best option in the irradiance and insolation forecasting.

\begin{landscape}
\fontsize{7}{11}\selectfont{
\begin{longtable}[c]{p{0.8cm}l|lll|llll|llll|llll}
\hline
 &  & \multicolumn{3}{|c|}{Complete} & \multicolumn{4}{c|}{MAE} & \multicolumn{4}{c|}{RMSE} & \multicolumn{4}{c}{MBE} \\ \cline{3-17}
\endhead
AWS & Model & MAE & MBE & RMSE & Q1 & Q2 & Q3 & Q4 & Q1 & Q2 & Q3 & Q4 & Q1 & Q2 & Q3 & Q4 \\ \hline
\parbox[t]{2mm}{\multirow{4}{*}{\rotatebox[origin=c]{90}{Biotopo}}}  & ARIMA & 0.061 &0.004 &0.081
 & 0.246	&\textbf{0.109}	&\textbf{0.156}	&0.072	&0.163 & \textbf{0.085}	&\textbf{0.103}	&0.093 &0.023	&\textbf{-0.050}&\textbf{-0.037}	&0.006
\\ \cline{2-17} 
 & SL-FNN & 0.062	&0.008	&0.089& 0.349&	0.943&	1.394&	0.062	&0.260	&0.672	&0.814	&0.080	&0.099	&0.942	&1.391	&0.001
 \\ \cline{2-17} 
 & LSTM & \textbf{0.050} &	\textbf{0.001}	&\textbf{0.067}
 & \textbf{0.205}&	0.669	&0.985	&\textbf{0.058}&	\textbf{0.144}&	0.490&	0.597&	\textbf{0.074}&	\textbf{0.015}	&0.656&	0.959&	0.000
 \\ \cline{2-17} 
 & ML-FNN & 0.054	&0.003	&0.075
 & 0.266	&0.726&	1.096&	0.061	&0.195	&0.519	&0.643	&0.078	&0.044	&0.722	&1.088	&\textbf{0.000}
 \\ \hline
\parbox[t]{2mm}{\multirow{4}{*}{\rotatebox[origin=c]{90}{Altaquer}}} & ARIMA & 0.087	&0.007	&0.112
 & 0.293	&0.230	&0.343	&0.100&	0.192&	0.179&	0.224&	0.129	&\textbf{0.023}&	-0.082&	\textbf{-0.034}&	0.008
 \\ \cline{2-17} 
 & SL-FNN & 0.180	&0.088	&0.234
 & 1.524&	0.909&	1.435&	0.112&	0.804&	0.660&	0.848&	0.133&	1.524&	0.909&	1.433&	-0.065
 \\ \cline{2-17} 
 & LSTM & \textbf{0.075}	&\textbf{0.002}	&\textbf{0.093}
 & \textbf{0.308}&	\textbf{0.192}&	\textbf{0.313}&	0.093&	\textbf{0.189}&	\textbf{0.157}&	\textbf{0.210}&	0.112&	0.031&	\textbf{0.016}&	0.107&	\textbf{-0.003}
 \\ \cline{2-17} 
 & ML-FNN & 0.162&	0.076&	0.219
 & 1.432&	0.942&	1.491&	\textbf{0.087}&	0.773&	0.686&	0.886&	\textbf{0.105}&	1.428&	0.942&	1.491&	0.035
 \\ \hline
\parbox[t]{2mm}{\multirow{4}{*}{\rotatebox[origin=c]{90}{\begin{tabular}[c]{@{}l@{}}Granja el \\ Mira\end{tabular}}}}  & ARIMA & 0.079	&0.007	&0.104
 & 0.297	&\textbf{0.079}&	\textbf{0.162}&	0.082&	0.189&	\textbf{0.087}&	\textbf{0.127}	&0.106&	\textbf{0.005}&	\textbf{0.027}&	\textbf{0.067}&	0.005
 \\ \cline{2-17} 
 & SL-FNN & 0.092&	-0.018&	0.119
 & 0.542&	0.612&	0.909&	0.070&	0.330&	0.450&	0.552&	0.088&	-0.441&	-0.612&	-0.905&	-0.007
 \\ \cline{2-17} 
 & LSTM & \textbf{0.066}&	\textbf{0.002}&	\textbf{0.082}
 & \textbf{0.282}&	0.195&	0.313&	0.065&	\textbf{0.167}&	0.151&	0.199&	0.081&	0.043&	0.184&	0.287&	\textbf{0.002}
 \\ \cline{2-17} 
 & ML-FNN & 0.077&	0.010&	0.097
 & 0.380&	0.375&	0.591&	\textbf{0.061}&	0.239&	0.278&	0.359	&\textbf{0.077}	&0.182&	0.375&	0.585&	0.019
 \\ \hline
 \parbox[t]{2mm}{\multirow{4}{*}{\rotatebox[origin=c]{90}{Guapi}}} & ARIMA & 0.062	&0.005&	0.086
 & 0.261&	0.157&	0.246&	0.079&	0.173&	0.130&	0.170&	0.104&	0.022&	\textbf{-0.023}&	\textbf{0.023}&	0.005
 \\ \cline{2-17} 
 & SL-FNN & 0.053	&0.000	&0.069
 & \textbf{0.224}	&\textbf{0.118}&	\textbf{0.183}&	0.059&	\textbf{0.145}&	\textbf{0.094}&	\textbf{0.122}&	0.077&	\textbf{0.002}&	-0.069&	-0.085&	\textbf{0.000}
 \\ \cline{2-17} 
 & LSTM & \textbf{0.050}	&\textbf{0.000}	&\textbf{0.068}
 & 0.229&	0.230&	0.326	&\textbf{0.058}&	0.151&	0.174&	0.207&	\textbf{0.076}&	0.004&	-0.197&	-0.259&	0.000
 \\ \cline{2-17} 
 & ML-FNN & 0.054&	0.001&	0.070 & 0.231&	0.379&	0.580&	0.059&	0.151&	0.278&	0.350&	0.077&	0.006&	0.371&	0.566&	0.001 \\ \hline
\parbox[t]{2mm}{\multirow{4}{*}{\rotatebox[origin=c]{90}{\begin{tabular}[c]{@{}l@{}}Viento \\ Libre \end{tabular}}}} & ARIMA & 0.075&	0.013&	0.100
 & 0.233&	0.803&	0.848&	0.080&	0.160&	0.759&	0.761&	0.105&	0.037&	-0.774&	-0.790&	0.013
 \\ \cline{2-17} 
 & SL-FNN & 0.071&	0.010&	0.097
 & 0.325&	1.162&	1.625&	\textbf{0.064}&	0.247&	0.846&	0.971&	\textbf{0.081}&	0.131&	1.162&	1.623&	0.002
 \\ \cline{2-17} 
 & LSTM & 0.062&	0.000&	0.081
 & 0.187&	\textbf{0.127}&	\textbf{0.178}&	0.067&	0.125&	\textbf{0.104}&	\textbf{0.126}&	0.084&	-0.004&	\textbf{0.064}&	\textbf{0.052}&	\textbf{0.000}
 \\ \cline{2-17} 
 & ML-FNN & \textbf{0.062}&	\textbf{0.000}&	\textbf{0.081}
 & \textbf{0.185}	&0.376&	0.516&	0.068&	\textbf{0.124}&	0.283&	0.323	&0.086&	\textbf{0.004}&	0.366&	0.497&	0.002
 \\ \hline
\parbox[t]{2mm}{\multirow{4}{*}{ \rotatebox[origin=c]{90}{\begin{tabular}[c]{@{}l@{}}Universidad \\ de Nariño\end{tabular}}}} & ARIMA & 0.080&	0.012&	0.105
 & 0.326&	0.337&	0.493&	0.083&	0.209&	0.253&	0.309&	0.104&	0.053&	0.242&	0.303&	0.011
 \\ \cline{2-17} 
 & SL-FNN & 0.068&	-0.001&	0.087
 & 0.310&	0.234&	0.329&	0.063&	0.194&	0.176&	0.206&	0.079&	-0.010&	-0.175&	-0.212&	\textbf{0.000}
 \\ \cline{2-17} 
 & LSTM & 0.072&	\textbf{0.000}&	0.114
 & 0.366&	0.462&	0.738&	0.065&	0.349&	0.403&	0.539&	0.080&	\textbf{0.002}&	0.343&	0.503&	0.000
 \\ \cline{2-17} 
 & ML-FNN & \textbf{0.068}	&0.000&	\textbf{0.085}
 & \textbf{0.273}&	\textbf{0.128}&	\textbf{0.231}&	\textbf{0.062}&	\textbf{0.170}&	\textbf{0.125}&	\textbf{0.175}&	\textbf{0.078}&	0.002&	\textbf{0.066}&	\textbf{0.107}&	0.000
 \\ \hline

 \parbox[t]{2mm}{\multirow{4}{*}{\rotatebox[origin=c]{90}{Josefina}}}  & ARIMA & 0.052&	0.005&	0.074
 & 0.200&	\textbf{0.133}&	\textbf{0.191}&	0.076&	0.142&	\textbf{0.110}&	\textbf{0.135}&	0.101&	0.015&	\textbf{0.082}&	\textbf{0.088}&	0.011
 \\ \cline{2-17} 
 & SL-FNN & 0.056&	0.011&	0.090
 & 0.352&	0.949&	1.425&	0.062&	0.289&	0.677&	0.832&	0.081&	0.150&	0.948&	1.423&	0.002
 \\ \cline{2-17} 
 & LSTM & \textbf{0.043}&	\textbf{0.000}&	\textbf{0.060}
 & \textbf{0.168}&	0.444&	0.652&	\textbf{0.060}&	\textbf{0.120}&	0.324&	0.394&	0.078&	\textbf{0.010}&	0.432&	0.628&	\textbf{0.000}
 \\ \cline{2-17} 
 & ML-FNN & 0.046&	0.001&	0.062
 & 0.177&	0.263&	0.385&	0.060&	0.125&	0.195&	0.236&	\textbf{0.078}&	0.012&	0.258&	0.375&	0.000
 \\ \hline
\parbox[t]{2mm}{\multirow{4}{*}{\rotatebox[origin=c]{90}{Botana}}} & ARIMA & 0.089&	0.012&	0.114
 & 0.364&	\textbf{0.212}&	\textbf{0.341}&	0.085&	0.229&	\textbf{0.180}&\textbf{	0.240}&	0.108&	0.052&	\textbf{0.152}&	\textbf{0.220}&	0.012
 \\ \cline{2-17} 
 & SL-FNN & 0.074 & -0.001&	0.093
 & 0.334	&0.259&	0.380&	0.065&	0.207&	0.206&	0.256&	0.082&	-0.014&	-0.255&	-0.371&	0.000
 \\ \cline{2-17} 
 & LSTM & 0.075&	\textbf{0.000}&	0.099
 & 0.350&	0.966&	1.415&	0.066&	0.241&	0.695&	0.835&	0.084&	\textbf{-0.002}&	-0.866&	-1.215&	\textbf{0.000}
 \\ \cline{2-17} 
 & ML-FNN & \textbf{0.072}&	0.002&	\textbf{0.091}
 & \textbf{0.328}&	0.471&	0.718&	\textbf{0.065}&	\textbf{0.205}&	0.343&	0.431&	\textbf{0.082}&	0.027&	0.468&	0.713&	0.001
 \\ \hline
\parbox[t]{2mm}{\multirow{4}{*}{\rotatebox[origin=c]{90}{Paraiso}}} & ARIMA & 0.072&	0.009&	0.096
 & 0.222&	\textbf{0.322}&	\textbf{0.382}&	0.084&	0.165&	\textbf{0.273}&	\textbf{0.282}&	0.106&	0.032&	\textbf{-0.232}&	\textbf{-0.199}&	0.012
 \\ \cline{2-17} 
 & SL-FNN & 0.061&	0.002&	0.081
 & 0.237&	0.525&	0.808&	\textbf{0.067}&	0.175&	0.380&	0.480&	0.085&	0.026&	0.522&	0.803&	-0.001
 \\ \cline{2-17} 
 & LSTM & \textbf{0.058}&	\textbf{0.001}&	\textbf{0.077}
 & \textbf{0.195}&	0.562&	0.830&	0.067&	\textbf{0.145}&	0.410&	0.501&	\textbf{0.084}&	\textbf{0.012}&	0.547&	0.801&	-0.001
 \\ \cline{2-17} 
 & ML-FNN & 0.060&	0.003&	0.079
 & \textbf{0.231}&	0.414&	0.625&	0.069&	0.162&	0.296&	0.367&	0.087&	0.044&	0.412&	0.621&	\textbf{0.000}
 \\ \hline
\parbox[t]{2mm}{\multirow{4}{*}{ \rotatebox[origin=c]{90}{\begin{tabular}[c]{@{}l@{}}Ospina \\ Perez \end{tabular}}}} & ARIMA & 0.051	&0.006&	0.085 & 0.334&	0.255&	0.417&	0.081&	0.222&	0.222&	0.301&	0.107&	0.037&	-0.131&	\textbf{-0.162}&	0.008
 \\ \cline{2-17} 
 & SL-FNN & \textbf{0.045}&	\textbf{-0.001}&	\textbf{0.069}
 & \textbf{0.315}&	\textbf{0.147}&	\textbf{0.254}&	\textbf{0.063}&	\textbf{0.198}&	\textbf{0.128}&	\textbf{0.176}&	\textbf{0.080}&	-0.024&	\textbf{0.106}&	0.178&	-0.009
 \\ \cline{2-17} 
 & LSTM & 0.067&	-0.002&	0.099
 & 0.500&	0.336&	0.505&	0.087&	0.302&	0.264&	0.334&	0.108&	\textbf{-0.024}&	-0.299&	-0.420&	\textbf{-0.003}
 \\ \cline{2-17} 
 & ML-FNN & 0.134&	0.065&	0.188
 & 1.269&	0.945&	1.414&	0.067&	0.688&	0.682&	0.837&	0.086&	1.269&	0.945&	1.413&	0.034
 \\ \hline
\parbox[t]{2mm}{\multirow{4}{*}{\rotatebox[origin=c]{90}{Sandona}}} & ARIMA & 0.062	&0.007	&0.084
 & 0.305&	0.577&	0.653&	0.073&	0.206&	0.505&	0.511&	0.096&	0.024&	-0.495&	-0.488&	0.012
 \\ \cline{2-17} 
 & SLP & 0.058&	-0.004&	0.075
 & \textbf{0.264}&	\textbf{0.203}&	\textbf{0.255}&	0.056&	\textbf{0.168}&	\textbf{0.167}&	\textbf{0.180}&	0.078&	\textbf{-0.023}&	\textbf{-0.124}&	\textbf{-0.095}&	-0.020
 \\ \cline{2-17} 
 & LSTM & \textbf{0.052}&	\textbf{0.005}&	\textbf{0.070}
 & 0.302&	0.239&	0.409&	\textbf{0.053}&	0.196&	0.198&	0.271&	0.073&	0.085&	0.232&	0.392&	\textbf{-0.005}
 \\ \cline{2-17} 
 & ML-FNN & 0.079&	0.021&	0.105
 & 0.556&	0.373&	0.612&	0.055&	0.337&	0.282&	0.374&	\textbf{0.071}&	0.473&	0.373&	0.609&	-0.007
 \\ \hline
\parbox[t]{2mm}{\multirow{4}{*}{ \rotatebox[origin=c]{90}{\begin{tabular}[c]{@{}l@{}}Cerro \\ Páramo\end{tabular}}}} & ARIMA & 0.074&	0.009&	0.105 & 0.297&	0.417&	0.509&	0.069&	0.203&	0.340&	0.356&	0.094&	0.034&	0.319&	\textbf{0.315}&	0.008
 \\ \cline{2-17} 
 & SL-FNN & \textbf{0.066}&	\textbf{-0.001}&	\textbf{0.087}
 & \textbf{0.278}&	\textbf{0.200}&	\textbf{0.343}&	\textbf{0.059}&	\textbf{0.186}&	\textbf{0.176}&	\textbf{0.242}&	\textbf{0.079}&	-0.014&	\textbf{-0.192}&	-0.328&	0.000
 \\ \cline{2-17} 
 & LSTM & 0.074&	0.001&	0.098
 & 0.312&	0.824&	1.186&	0.066&	0.210&	0.608&	0.725&	0.087&	\textbf{0.013}&	0.807&	1.152&	\textbf{0.000}
 \\ \cline{2-17} 
 & ML-FNN & 0.071&	0.003&	0.094
 & 0.343&	0.761&	1.142&	0.061&	0.230&	0.548&	0.675&	0.080&	0.024&	0.755&	1.130&	0.002
 \\ \hline
\caption{Insolation statistical errors of the one day-ahead forecasting process}
\label{tab:d_err}\\
\end{longtable}
}
\end{landscape}

\section{Conclusions}\label{S5}

In this work, we implemented four forecasting models: ARIMA, SL-FNN, ML-FNN, and LSTM for forecasting global solar irradiance with one-day ahead horizons in an hourly timestamp and global solar insolation with one-day ahead horizons in an daily timestamp. One of the challenges we tackled was the handling of missing data. We implemented an imputation process to deal with this issue. The training data preprocessing required the imputation of more than 50 \% of the time series values. \textit{Altaquer}, that provided the most complete time series, required the imputation of 54,7 \% of the data, and \textit{La Josefina} was the most critical case and required the imputation of 84,2 \% of the data. The statistical errors of the solar irradiance imputation showed that the resource was underestimated in most cases. The RMSE and MAE result showed that in the Andean zone the error is larger than in the Pacific zone. For the solar irradiance imputation, we use the empirical temperature-based models Hargreaves and Samani and Logistic. The former for the Pacific zone and the latter for the Andean and Amazonia zones. To reduce the statistical errors, it would be worth exploring alternative imputation techniques for solar irradiance and insolation data. 

In the hourly forecasting, the SL-FNN, ML-FNN and LSTM outperformed ARIMA in most cases. In Pacific zone, LSTM was the best model when training with the shortest time series (\textit{Altaquer}), and SLP when training with longer ones (\textit{Biotopo} and \textit{Guapi}). Also, the neural network-based models showed a better performance in cloudy conditions in this zone (\textit{Biotopo} and \textit{Altaquer}). This result indicates that the solar irradiance randomness induced by the cloud motion might be better modeled by neural network-based models.

In the Andean zone, SL-FNN presents better results in the MAE and RMSE error measurement when the time series is longer i.e. in \textit{Viento Libre}, \textit{Botana}, \textit{La Josefina}, \textit{Paraiso}, and \textit{Universidad de Nariño} cases. When the time series is shorter (\textit{Ospina Perez} and \textit{Sandona}), ARIMA is the model with the smallest RMSE and MAE error. Therefore, for the Andean zone one year of measurements on average was not enough for training the neural network-based models. As a result, in such cases ARIMA is a better option. The MBE error measurement shows LSTM as the model with the smallest bias in all AWS of this zone. In the Amazonian zone, the \textit{Cerro Páramo} AWS,  SL-FNN is the best model in all error measurements. Overall, the SL-FNN model outperforms the other models in global solar irradiance in a one-day ahead horizon with an hourly frequency. As a conclusion, SL-FNN is the best option for sites with high altitude and cloudiness. 

The statistical error measurements of the global solar insolation forecasting in a daily timestamp show that LSTM is the best option in the Pacific zone regardless the time series length. In the Andean zone, the neural network-based models exhibit lower statistical errors. However, the model performance depends on the amount of training data. For example in the shortest time series LSTM and SL-FNN are the better models, while ML-FNN is the model with the lowest performance. The results presented in this study do not exhibit a clear pattern that indicates that there is one forecasting technique that outperforms the others in the Andean zone. As observed in other studies \cite{Dannecker2015}, our results indicate that the performance of each forecasting model depends on the specific task.

\section*{Acknowledgements}
The authors would like to thank the \textit{Fundación Ceiba} for their support in funding the doctoral studies of Laura Sofia Hoyos-Gómez.

\bibliographystyle{elsarticle-num}

\bibliography{cas-sc-ref}





\end{document}